\documentclass[10pt,journal,compsoc]{IEEEtran}
\usepackage{cite}
\usepackage{algorithmic}
\usepackage{array}
\usepackage{amssymb}
\usepackage{eqparbox}
\usepackage{url}
\usepackage{multirow}
\usepackage[cmex10]{amsmath}
\usepackage{makeidx}
\usepackage{diagbox}
\usepackage{rotating,soul}
\usepackage{wrapfig}
\usepackage{booktabs}
\usepackage[usenames,dvipsnames]{color}
\usepackage{ragged2e}
\usepackage{epsfig}
\usepackage{caption}
\usepackage{overpic}
\usepackage{enumitem}
\usepackage[dvipsnames,table,xcdraw,HTML]{xcolor}
\usepackage[linesnumbered,ruled]{algorithm2e}
\usepackage{colortbl}
\usepackage{graphicx}
\usepackage{contour}
\usepackage{listings}
\usepackage{subfig}  % \subfloat
\usepackage{amssymb}
\usepackage{pifont}
\usepackage{bm}
\usepackage{amssymb}
\usepackage{bbding}
\usepackage{wasysym}
\usepackage[pdfborder=false,colorlinks,bookmarksnumbered,bookmarksopen,linkcolor=black,citecolor=black,urlcolor=black]{hyperref}
\usepackage{orcidlink}

\makeatletter
  \newcommand\figcaption{\def\@captype{figure}\caption}
  \newcommand\tabcaption{\def\@captype{table}\caption}
\makeatother

\newcommand{\myparagraph}[1]{\textbf{#1}\hspace{1.8ex}}

\newcommand{\pub}[1]{\color{gray}{\tiny{#1}}}
\newcommand{\Frst}[1]{{\textbf{#1}}}
\newcommand{\Scnd}[1]{{{#1}}}

\newcommand{\addFig}[1]{}
\newcommand{\addFigs}[1]{}

\newcolumntype{x}[1]{>{\centering\arraybackslash\hspace{0pt}}p{#1}}

\newtheorem{myDef}{Definition}

\makeindex
\definecolor{cred}{HTML}{FF6B6B}
\definecolor{cyellow}{HTML}{FEC260}
\definecolor{cgreen}{HTML}{70AD47}
\definecolor{cblue}{HTML}{4D96FF}
\definecolor{cpurple}{HTML}{2A0944}
\definecolor{ggray}{RGB}{127,127,127}
\definecolor{aliceblue}{rgb}{0.94, 0.97, 1.0}
\definecolor{mygreen}{RGB}{0,150,0}
\definecolor{myred}{RGB}{200,0,0}

\DeclareGraphicsExtensions{.jpg,.pdf,.png}

\def\ie{{\em i.e.,~}}

\definecolor{mygreen}{RGB}{0,100,0}
\definecolor{myblue}{RGB}{10,100,200}
\definecolor{myred}{RGB}{200,0,0}

\begin{document}

%%%%%%%%% TITLE
\title{Hierarchical Banzhaf Interaction for General Video-Language Representation Learning}

\author{Peng Jin~\orcidlink{0000-0001-9287-6410}, Hao Li~\orcidlink{0000-0002-3200-0270}, Li Yuan~\orcidlink{0000-0002-2120-5588}, Shuicheng Yan~\orcidlink{0000-0001-8906-3777},~\IEEEmembership{Fellow,~IEEE}, and Jie Chen~\orcidlink{0000-0002-9765-4523}
  \IEEEcompsocitemizethanks{
    \IEEEcompsocthanksitem Peng Jin and Hao Li contributed equally to this work.
    \IEEEcompsocthanksitem Peng Jin and Hao Li are with the School of Electronic and Computer Engineering, Peking University, Shenzhen, China, and also with the Pengcheng Laboratory, Shenzhen, China. E-mail: {jp21@stu.pku.edu.cn, lihao1984@pku.edu.cn}.
    \IEEEcompsocthanksitem Li Yuan and Jie Chen are with the School of Electronic and Computer Engineering, Peking University, Shenzhen, China, and also with the Pengcheng Laboratory, Shenzhen, China. E-mail: {yuanli-ece@pku.edu.cn, chenj@pcl.ac.cn}.
    \IEEEcompsocthanksitem Shuicheng Yan is with Skywork AI and Nanyang Technological University. E-mail: shuicheng.yan@gmail.com.
    \IEEEcompsocthanksitem Corresponding authors: Li Yuan, Jie Chen.
	}
}

\markboth{IEEE TRANSACTIONS ON PATTERN ANALYSIS AND MACHINE INTELLIGENCE,~Vol.~xx,No.~xx,~xxx.~xxxx}%
{Jin \MakeLowercase{\emph{et al.}}: Hierarchical Banzhaf Interaction for General Multimodal Representation Learning}
%\pubid{0000--0000/00\$00.00~\copyright~2009 IEEE}

\IEEEcompsoctitleabstractindextext{%
\begin{abstract}
\justifying
Multimodal representation learning, with contrastive learning, plays an important role in the artificial intelligence domain. As an important subfield, video-language representation learning focuses on learning representations using global semantic interactions between pre-defined video-text pairs. However, to enhance and refine such coarse-grained global interactions, more detailed interactions are necessary for fine-grained multimodal learning. In this study, we introduce a new approach that models video-text as game players using multivariate cooperative game theory to handle uncertainty during fine-grained semantic interactions with diverse granularity, flexible combination, and vague intensity. Specifically, we design the Hierarchical Banzhaf Interaction to simulate the fine-grained correspondence between video clips and textual words from hierarchical perspectives. Furthermore, to mitigate the bias in calculations within Banzhaf Interaction, we propose reconstructing the representation through a fusion of single-modal and cross-modal components. This reconstructed representation ensures fine granularity comparable to that of the single-modal representation, while also preserving the adaptive encoding characteristics of cross-modal representation. Additionally, we extend our original structure into a flexible encoder-decoder framework, enabling the model to adapt to various downstream tasks. Extensive experiments on commonly used text-video retrieval, video-question answering, and video captioning benchmarks, with superior performance, validate the effectiveness and generalization of our method. The code is available at \hyperlink{https://github.com/jpthu17/HBI}{https://github.com/jpthu17/HBI}.
\end{abstract}

}

\maketitle
\IEEEdisplaynontitleabstractindextext
\IEEEpeerreviewmaketitle

\IEEEraisesectionheading{\section{Introduction}\label{sec:introduction}}

%%%%%%%%% BODY TEXT
\IEEEPARstart{M}ultimodal representation learning, which aims to narrow the gap among different modalities, is crucial for making the most of multimodal data~\cite{cheng2024parallel}. As an essential subfield, video-language representation learning aims to understand the relationship between videos and their associated textual descriptions. It is beneficial for various downstream tasks including text-video retrieval, video-question answering~(VideoQA), and video captioning. Recently, an increasing number of video-language representation learning models~\cite{li2022align,jin2022expectation,li2023freestyleret} apply contrastive learning to project the video and text features into a common latent space based on the semantic similarities of video-text pairs. In this manner, multimodal contrastive learning enables networks to learn discriminative video-language representations.

\begin{figure}[tbp]
\centering
\includegraphics[width=1.0\linewidth]{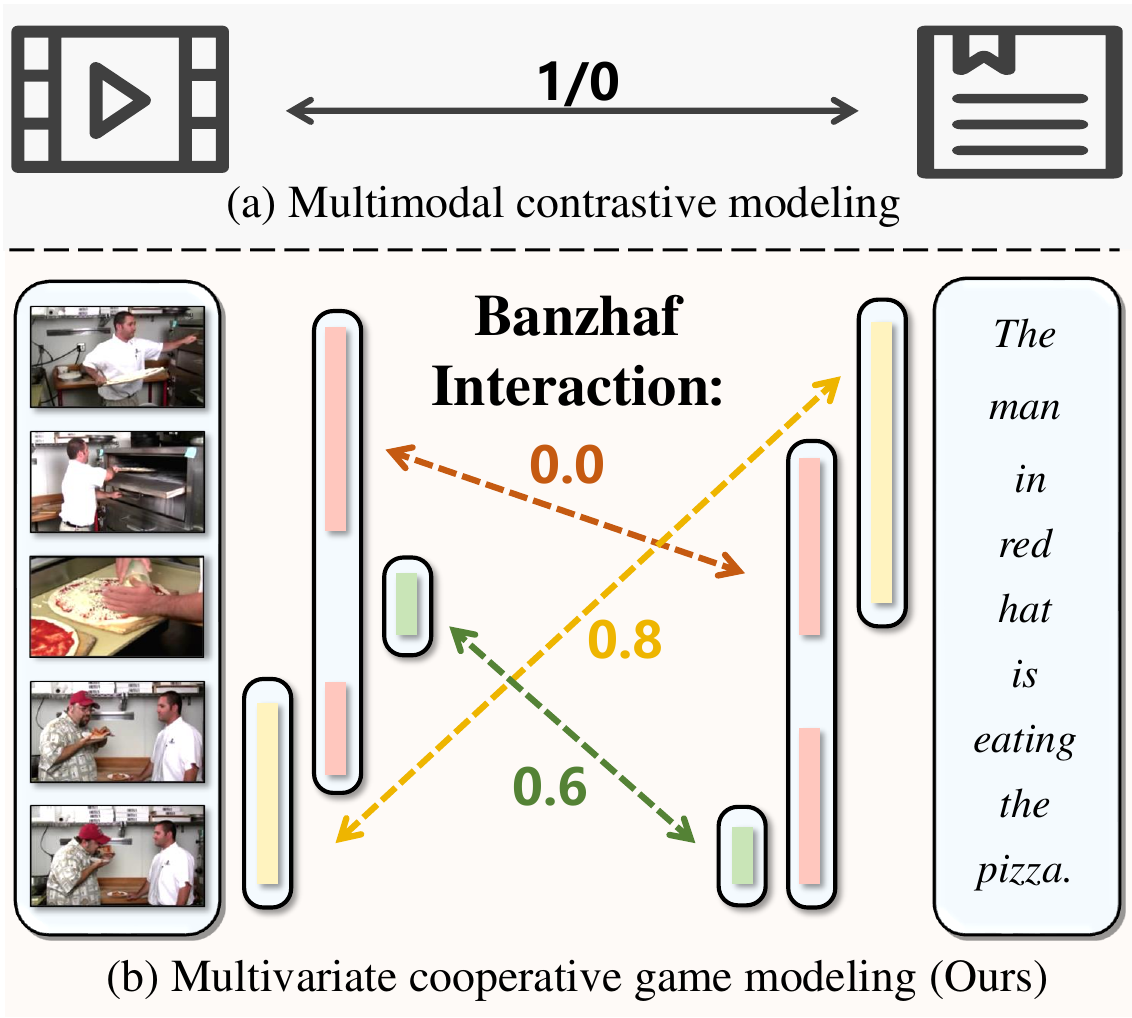}
\vspace{-0.2in}
\caption{
(a) Previous methods only learn a global semantic interaction from the coarse-grained labels of video-text pairs. (b) We model multimodal alignment as a cooperative game process, utilizing Banzhaf Interaction to evaluate possible correspondence between video frames and text words.
}
\label{intro: fig1}
\vspace{-0.12in}
\end{figure}

\begin{figure*}
\centering
\includegraphics[width=1.\linewidth]{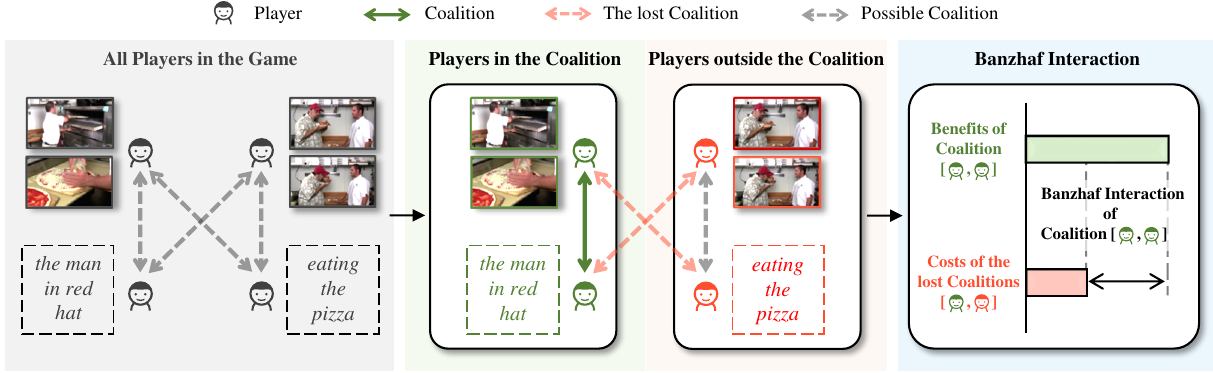}
\vspace{-0.2in}
\caption{\textbf{The intuition of employing Banzhaf Interaction in video-language representation learning.} When certain players (frames and words) form a coalition, it entails the exclusion of these players from potential coalitions with others, rendering them mutually exclusive from the target coalition. Banzhaf Interaction quantifies the disparity between the benefits derived from the coalition and the costs incurred due to the lost coalitions. Therefore, Banzhaf Interaction effectively captures the incremental benefits conferred by the coalition. We refer the reader to Eq.~\ref{BI} for the detailed formula.}
\label{intro: fig2}
\vspace{-0.12in}
\end{figure*}

Lacking fine-grained alignment annotations, the video-language contrastive learning models~\cite{wang2022many,dong2021dual,han2022temporal,jin2023diffusionret} typically perform coarse-grained feature alignment based on the global similarity between the video and the text. As shown in Fig.~\ref{intro: fig1}(a), previous contrastive models only exploit the coarse-grained labels of video-text pairs to learn a global semantic interaction. However, for multimodal downstream tasks like text-video retrieval and VideoQA, models must possess the capability to capture detailed and interpretable features. The coarse-grained text-video alignment hinders the interaction process between the visual entity and the textual phrase. Establishing fine-grained alignment through naive methods would require high-quality annotated data to explore clear video-text relationships, which is currently unavailable, particularly on large-scale vision-language datasets. This underscores the possibility of other learning signals to improve typical contrastive learning.

To address the above fine-grained alignment problem, we propose an innovative approach to model video-language representation learning as a multivariate cooperative game by formulating video and text as players in a cooperative game, as shown in Fig.~\ref{intro: fig1}(b). Intuitively, when visual and textual representations exhibit high semantic similarity, they are more likely to collaborate and enhance the overall multimodal similarity score. Guided by this insight, we regard the collection of multiple representations as a coalition and suggest using the game-theoretic interaction index, \ie Banzhaf Interaction~\cite{grabisch1999axiomatic}, to measure the degree of cooperation within a coalition. The Banzhaf Interaction~\cite{marichal2011weighted} is a well-known concept in cooperative game theory. As shown in Fig.~\ref{intro: fig2}, it quantifies the additional benefits accruing to a coalition compared to the costs of lost coalitions with other players. A high Banzhaf Interaction of a coalition indicates its greater contribution to the semantic similarity. Hence, it can help evaluate the potential correspondence between video frames and text words, making cross-modal contrastive learning sensitive and explainable.

In our preliminary conference paper~\cite{HBI}, we propose Hierarchical Banzhaf Interaction~(HBI), which treats video frames and text words as players, leveraging cross-modality similarity measurement as the characteristic function in the cooperative game. Specifically, HBI utilizes Banzhaf Interaction to delineate the trend of cooperation among any set of features. Moreover, to efficiently establish coalitions among game players, HBI introduces a token merge module to cluster the original frames (words) and reduce the number of players. By stacking token merge modules, HBI achieves hierarchical interaction, encompassing entity-level interactions among frames and words, action-level interactions among clips and phrases, and event-level interactions among segments and paragraphs.

\begin{figure*}
\centering
\includegraphics[width=1.\linewidth]{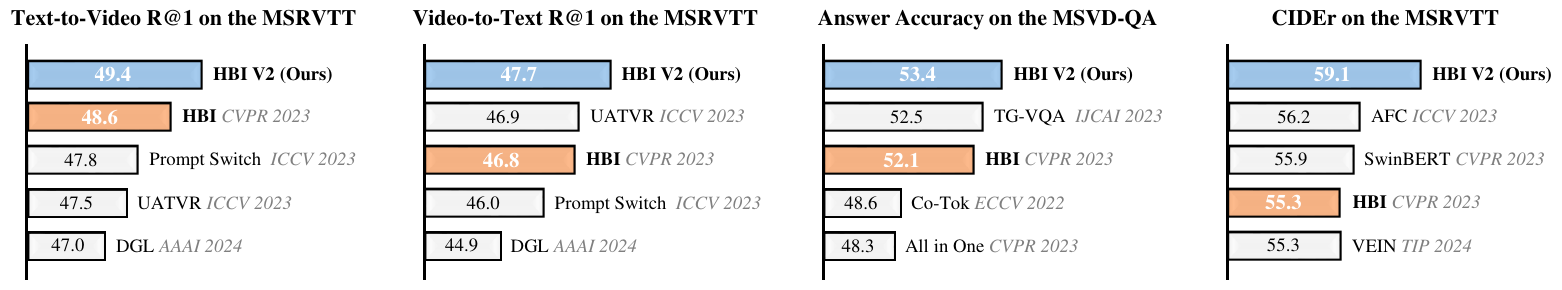}
\vspace{-0.2in}
\caption{\textbf{Performance comparisons on text-video retrieval, video-question answering, and video captioning.} Our proposed framework, HBI V2, designed for general video-language representation learning, demonstrates superior performance consistently. Notably, HBI V2 not only surpasses the previous HBI, but also outperforms existing task-specific methods.}
\label{intro: fig3}
\vspace{-0.12in}
\end{figure*}

Although our HBI framework has shown noticeable improvements, its current approach to game interactions solely relies on independently encoded video and text representations. This makes it hard to accurately calculate Banzhaf Interaction. Specifically, videos may contain redundant information, as illustrated in Fig.~\ref{intro: fig2}, which can have a bias in the calculations for Banzhaf Interaction. Employing cross-modal joint encoding helps select important video content based on the query text, reducing the bias in Banzhaf Interaction. However, since the cross-modal representation can only encode the entire video and text, it cannot be directly applied to our fine-grained game interaction modeling. What is worse, when semantic alignment between text and video is lacking, this text-conditioned representation might not only fail to enhance single-modal representations but may also perform worse than single-modal representations. 

To this end, we significantly extend our previously proposed HBI framework to introduce HBI V2, which combines the strengths of both single-modal and cross-modal representations. In HBI V2, we reconstruct the representation by integrating single-modal and cross-modal components. This reconstructed representation dynamically adjusts the weights of single-modal and cross-modal components, ensuring fine granularity comparable to that of the single-modal representation to appropriate our aim of fine-grained game interaction modeling while preserving the adaptability of cross-modal representation.

Furthermore, our previously proposed HBI framework primarily addresses one task, \ie text-video retrieval. To enhance the versatility of HBI V2 across various downstream tasks, including VideoQA and video captioning, we expand our original structure into a flexible encoder-decoder framework comprising an encoder and a task-specific decoder. We conduct comprehensive experiments on multiple widely used datasets, including three text-video retrieval datasets (\textit{MSRVTT}~\cite{xu2016msr}, \textit{ActivityNet Captions}~\cite{krishna2017dense}, \textit{DiDeMo}~\cite{anne2017localizing}), three VideoQA datasets (\textit{MSRVTT-QA}~\cite{xu2017video}, \textit{MSVD-QA}~\cite{xu2017video}, \textit{ActivityNet-QA}~\cite{yu2019activitynet}), and one video captioning dataset (\textit{MSRVTT}~\cite{xu2016msr}). As shown in Fig.~\ref{intro: fig3}, our proposed HBI V2 consistently surpasses both the previous HBI and existing methods across all downstream tasks. These results demonstrate the effectiveness of our proposed HBI V2 framework. Our main contributions are as follows:

\begin{itemize}
    \item To the best of our knowledge, we are the first to introduce the multivariate cooperative game process into fine-grained video-language learning. 
        
    \item To mitigate the bias in Banzhaf Interaction, we propose to reconstruct the representation as a fusion of single-modal and cross-modal components. This reconstructed representation retains the advantages of both fine-grained and adaptive encoding.
  
    \item With task-specific prediction heads, HBI V2 achieves impressive performance on various tasks of video-language learning, including text-video retrieval, video question answering, and video captioning.
    
\end{itemize}

%-------------------------------------------------------------------------
\section{Related Work}

\subsection{Visual-Language Learning}
Recently, contrastive learning methods show great success in cross-modal tasks~\cite{wei2021universal,li2023weakly,buch2022revisiting,li2022fine,jin2024chat,zhu2023languagebind,li2024decoupled,xu2024llava,yu2024evagaussians,tang2024cycle3d,zhang2025repaint123,tang2024cycle3d,pang2024next}, such as text-video retrieval, VideoQA, and video captioning. The main challenge of cross-modal learning is to use vision-language pairs to learn common representations shared between modalities. 

\myparagraph{Text-Video Retrieval.}
Text-video retrieval requires models to establish the correct match between texts and videos. Most works~\cite{chen2021learning,jin2023diffusionret,yang2024dgl} of text-video retrieval map text and video to the same semantic space, where the similarity between text features and video features can be directly calculated. Recently, contrastive learning methods have shown great success in advancing the state-of-the-art performance of the retrieval task. Contrastive learning methods~\cite{radford2021learning,jin2022expectation,cheng2022vista} try to learn data representations from positive and negative pairs, making the representations of positive pairs have high similarity, and negative pairs have low similarity. Due to manually labeling the fine-grained relationships being unavailable, contrastive learning cannot capture fine-grained information in a supervised manner. To this end, we model video-text as game players with cooperative game theory and propose to combine Banzhaf Interaction with multimodal representation learning.

\myparagraph{Video Question Answering.}
VideoQA~\cite{xu2021sutd,le2020hierarchical,zhong2022video} requires models to analyze the complex semantic correlation between the video and the question. Recently, several contrastive learning-based VideoQA models~\cite{lei2021less,piergiovanni2022video} use the contrastive loss for cross-modality explicit alignment and fusion. With the contrastive loss, VideoQA models successfully map the different modalities into the same latent space. However, VideoQA requires element-level matching between text entities and video clips to predict the correct answer. Lacking the element-level fine-grained matching annotations, existing contrastive-learning-based VideoQA models suffer from slow convergence, requiring massive data~\cite{bain2021frozen,huang2021multilingual} for pretraining. In contrast to prior works, we explicitly capture the fine-grained semantic relationships between video and text via Banzhaf Interaction.

\myparagraph{Video Captioning.}
Video captioning requires models to automatically describe videos using natural language sentences. Existing methods~\cite{li2018jointly,pan2017video} adopt the encoder-decoder architecture to generate captions flexibly from the video features. Recently, contrastive-learning-based video captioning models~\cite{pan2016jointly,chen2023retrieval} align the video clips and textual entities to enhance the quality of the generated video descriptions. However, similar to retrieval and VideoQA, existing contrastive-learning-based models fail to achieve element-level concept matching, restricting the caption quality. In this work, we demonstrate that the proposed HBI V2 also achieves competitive performance for video captioning.

\begin{figure*}[htbp]
    \centering
    \includegraphics[width=1.\linewidth]{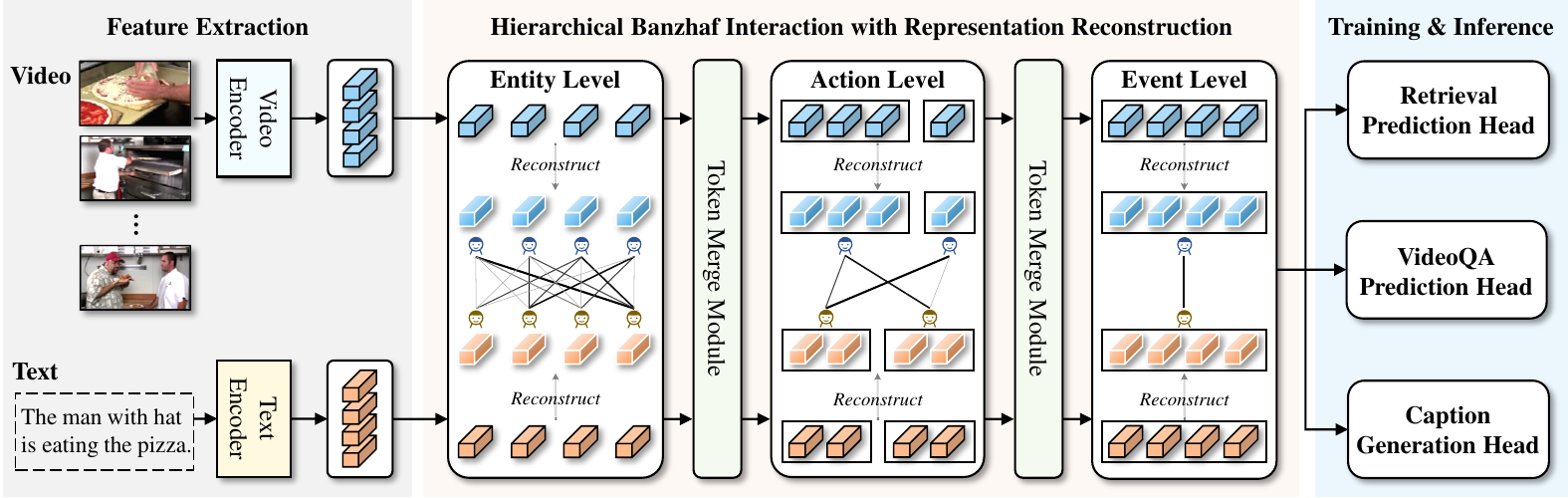}
    \vspace{-0.2in}
    \caption{
    \textbf{Overview of our proposed HBI V2 framework.} We employ a dual-stream encoder to extract features for video tokens and text tokens. Subsequently, we reconstruct the original representation by merging single-modal and cross-modal components. We propose a novel proxy training objective, which uses Banzhaf Interaction to evaluate possible correspondence between video tokens and text tokens from various levels. Furthermore, we customize different task-specific prediction heads for various downstream tasks.}
    \label{method: fig1}
    \vspace{-0.12in}
\end{figure*}

\subsection{Cooperative Game Theory} 
The game-theoretic interaction~\cite{ferguson2020course} consists of a set of players with a characteristic function. The characteristic function maps each team of players to a real number which indicates the payoff obtained by all players working together to complete the task. The core of game-theoretic interaction is to allocate different payoffs to game individuals fairly and reasonably. There are several interaction strategies including core interaction~\cite{jeukenne1977optical}, Shapley interaction~\cite{sun2020random}, and Banzhaf interaction~\cite{marichal2011weighted}. The game-theoretic interaction has multiple applications in different fields~\cite{aflalo2022vl,datta2016algorithmic}.

Toward the goal of achieving fine-grained alignment between video and text, we propose to leverage multi-player game theory to construct an alignment label generator. We aim to obtain the semantic relationship between visual tokens and text tokens. And the game theory targets generating an appropriate coalition construction strategy for multiple players. Thus, we propose to introduce game theory to help the alignment label generation by considering the video clips and the text entities as players. 

The cooperative game theory consists of a set $\mathcal{N}=\{1,2,...,n\}$ of players with a characteristic function $\phi$. The characteristic function $\phi$ maps each team of players to a real number. This number indicates the payoff obtained by all players working together to complete the task. The core of the cooperative game theory is calculating how much gain is obtained and how to distribute the total gain fairly~\cite{sun2020random}.

In a cooperative game, some players tend to form a coalition: it may happen that $\phi(\{i\})$ and $\phi(\{j\})$ are small, and at the same time $\phi(\{i,j\})$ is large. The Banzhaf Interaction~\cite{grabisch1999axiomatic} measures the additional benefits brought by the target coalition compared with the costs of the lost coalitions of these players with others. The costs of the lost coalitions can be estimated by each player in the target coalition working individually. For a coalition $\{i,j\}$, we consider $[\{i,j\}]$ as a single hypothetical player, which is the union of the players in $\{i,j\}$. Subsequently, the reduced cooperative game is formed by removing the individual players in $\{i,j\}$ from the game and adding $[\{i,j\}]$ to the game.

\begin{myDef}
\textbf{Banzhaf Interaction.} Given a coalition $\{i,j\} \subseteq \mathcal{N}$, the Banzhaf Interaction $\mathcal{I}([\{i,j\}])$ for the player $[\{i,j\}]$ is defined as:
\begin{equation}
\begin{aligned}
\mathcal{I}([\{i,j\}])=\!\sum_{\!\mathcal{C} \subseteq \mathcal{N} \setminus \{i,j\} }p(\mathcal{C}){\Bigl[}\phi(\mathcal{C}\cup \{[\{i,j\}]\})+\\ \phi(\mathcal{C})-\phi(\mathcal{C}\cup\{i\})-\phi(\mathcal{C}\cup\{j\}){\Bigr]},
\end{aligned}
\label{BI}
\end{equation}
where $p(\mathcal{C})=\frac{1}{2^{n-2}}$ is the likelihood of $\mathcal{C}$ being sampled. ``$\mathcal{N} \setminus \{i,j\} $'' denotes removing $\{i,j\}$ from $\mathcal{N}$.
\end{myDef}
Intuitively, $\mathcal{I}([\{i,j\}])$ reflects the tendency of interactions inside $\{i,j\}$. The higher value of $\mathcal{I}([\{i,j\}])$ indicates that player $i$ and player $j$ cooperate closely with each other.

\begin{figure*}[htbp]
    \centering
    \includegraphics[width=1.\linewidth]{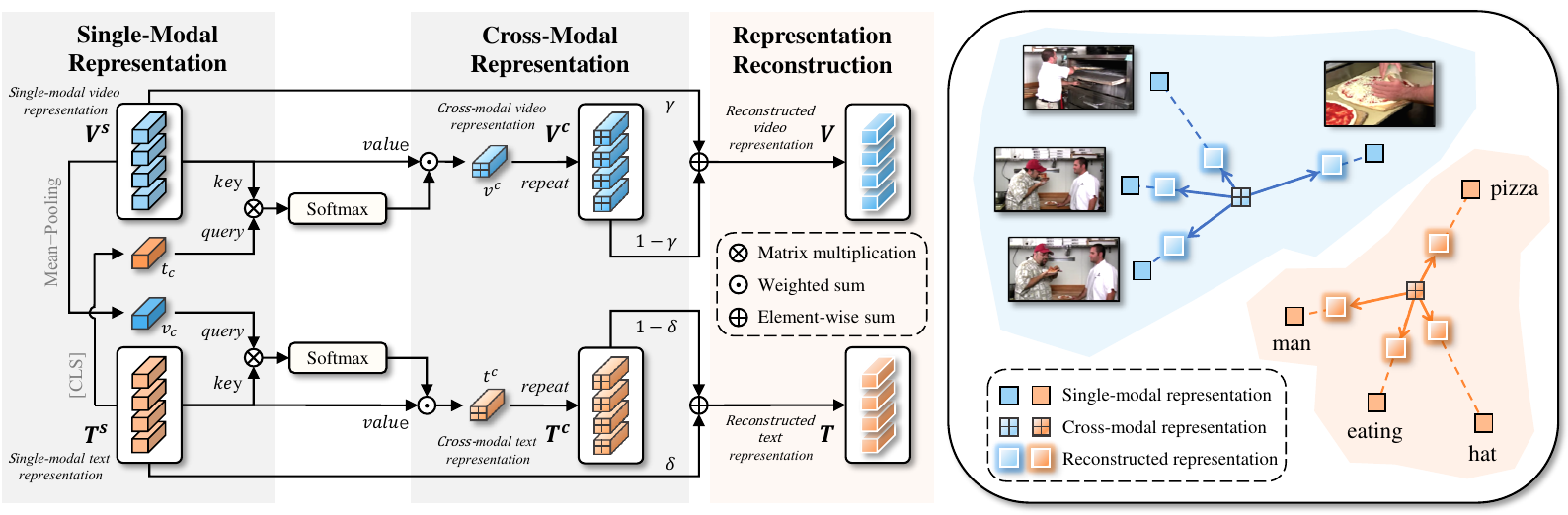}
    \vspace{-0.2in}
    \caption{
    \textbf{{The representation reconstruction module.}} To address the bias in calculations within Banzhaf Interaction, we reconstruct both video and text representation as a fusion of single-modal and cross-modal components. The representation reconstruction module maintains the granularity inherent in single-modal representations while preserving the adaptive encoding capabilities of cross-modal representations.}
    \label{method: fig2}
    \vspace{-0.12in}
\end{figure*}

\section{Methodology}
Our model consists of three submodules: (1)~\textbf{Representation Reconstruction}, which merges single-modal and cross-modal representations. (2)~\textbf{Hierarchical Banzhaf Interaction Module}, establishing multi-level Banzhaf Interaction between video and text by modeling them as game players. (3)~\textbf{Task-Specific Prediction Heads}, designed for various tasks. Fig.~\ref{method: fig1} shows the overview of our proposed HBI V2.

% Representation space optimization

\subsection{Representation Reconstruction}
Following previous works~\cite{jin2023diffusionret,jin2023text}, we employ CLIP~\cite{radford2021learning} as the backbone for generating both visual and textural representations. Specifically, for video representation, we evenly sample frames from the video to form the input frame sequence. Subsequently, we leverage ViT~\cite{dosovitskiy2021an} to encode the frame sequence and adapt the output from the [CLS] token as the frame embedding. Following this, we employ a 4-layer transformer to aggregate the embedding of all frames and obtain the frame representation ${\bm{V}}_{f}^{s} = \{\hat{v}^{i}_{f}\}_{i=1}^{N_v}$, where $\hat{v}^{i}_{f}$ signifies a frame feature and $N_v$ denotes the total number of frames. For text representation, we utilize the text encoder of CLIP to obtain the text representation ${\bm{T}}_{w}^{s}=\{\hat{t}^{j}_{w}\}_{j=1}^{N_t}$, where $N_t$ indicates the length of text tokens. 

Dynamically encoding video based on query text helps reduce bias in Banzhaf Interaction. However, since the cross-modal representation can only encode the entire video and text, it cannot be directly applied to our fine-grained game interaction modeling. When the semantic content of the text and video diverge greatly, this text-conditioned representation might not only fail to enhance single-modal representations but also might potentially perform worse than original single-modal representations. To this end, we propose to reconstruct the representation as a fusion of single-modal and cross-modal components. This reconstructed representation offers the combined benefits of the granularity inherent in single-modal representations and the adaptability of cross-modal representations. As shown in Fig.~\ref{method: fig2}, we apply this representation reconstruction method to both video and text, enabling the enhanced encoding of both modalities.

Starting with the video representation, we first aggregate frame representation with the text-frame attention encoder. Concretely, we feed the [CLS] token ${t}_{c}$ from the text representation as a query and the single-modal representation ${\bm{V}}_{f}^{s}$ as keys and values into an attention module. The resulting cross-modal video representation ${\bm{V}}_{f}^{c}$ is defined as:
\begin{equation}
\begin{aligned}
{\bm{V}}_{f}^{c}=\{{v}^{c}\}^{N_v}, \quad {v}^{c}=\textrm{Softmax}({t}_{c} {\bm{V}_{f}^{s}}^\top){\bm{V}}_{f}^{s}.
\end{aligned}
\end{equation}
Subsequently, the video representation ${\bm{V}}_{f}$ is defined as:
\begin{equation}
{\bm{V}}_{f}=\gamma{\bm{V}}_{f}^{s}+ (1-\gamma){\bm{V}}_{f}^{c},
\end{equation}
where $\gamma=\textrm{MLP}({\bm{V}}_{f}^{s}-{\bm{V}}_{f}^{c})$ serves as a factor used to adjust the balance between the single-modal and cross-modal components. When $\gamma=0$, the video representation is equivalent to the cross-modal video representation, \ie ${\bm{V}}_{f}={\bm{V}}_{f}^{c}$. Conversely, when $\gamma=1$, the video representation corresponds to the single-modal video representation, \ie ${\bm{V}}_{f}={\bm{V}}_{f}^{s}$.

To reconstruct the text representation, we employ the mean of the video frame representation, denoted as ${v}_{c}=\textrm{MeanPool}({\bm{V}}_{f}^{s})$, as a query. Subsequently, we use this query along with the single-modal representation ${\bm{T}}_{w}^{s}$ as keys and values to obtain the cross-modal text representation ${\bm{T}}_{w}^{c}$:
\begin{equation}
\begin{aligned}
{\bm{T}}_{w}^{c}=\{{t}^{c}\}^{N_t}, \quad {t}^{c}=\textrm{Softmax}({v}_{c} {\bm{T}_{w}^{s}}^\top){\bm{T}}_{w}^{s}.
\end{aligned}
\end{equation}
Finally, the text representation ${\bm{T}}_{w}$ is defined as:
\begin{equation}
{\bm{T}}_{w}=\delta{\bm{T}}_{w}^{s}+(1-\delta){\bm{T}}_{w}^{c},
\end{equation}
where $\delta=\textrm{MLP}({\bm{T}}_{w}^{s}-{\bm{T}}_{w}^{c})$ serves as a factor to adjust the proportion of single-modal and cross-modal components. When $\delta=0$, the text representation is equivalent to the cross-modal text representation, \ie ${\bm{T}}_{w}={\bm{T}}_{w}^{c}$. Conversely, when $\delta=1$, the text representation corresponds to the single-modal text representation, \ie ${\bm{T}}_{w}={\bm{T}}_{w}^{s}$.

\subsection{Hierarchical Banzhaf Interaction}
Generally, given a corpus of video-text pairs $(\bm{v},\bm{t})$, cross-modal representation learning aims to learn a video encoder and a text encoder. The problem is formulated as a cross-modality similarity measurement $\textrm{S}_{\bm{v},\bm{t}}$ by cross-modal contrastive learning, where the matched video-text pairs are close and the mismatched pairs are away from each other.

As the fine-grained relationships are unavailable, prior works typically directly apply the cross-modal contrastive loss to optimize the similarity scores $\textrm{S}_{\bm{v},\bm{t}}$. To move a step further, we model video-text as game players with multivariate cooperative game theory to handle the uncertainty during fine-grained semantic interaction with diverse granularity, flexible combination, and vague intensity.

\subsubsection{Video-Text as Game Players}
Given features $\bm{V}_{f}=\{v^{i}_{f}\}^{N_v}_{i=1}$ and $\bm{T}_{w}=\{t^{j}_{w}\}^{N_t}_{j=1}$, fine-grained cross-modal learning aims to find semantically matched video-text feature pairs. Specifically, if a video frame and a text word have strong semantic correspondence, then they tend to cooperate with each other and contribute to the fine-grained similarity score. Thus, we can consider $\mathcal{N}=\{v^{i}_{f}\}^{N_v}_{i=1}\cup \{t^{j}_{w}\}^{N_t}_{j=1}$ as players in the game. 

To achieve the goal of the cooperative game and cross-modal learning to be completely consistent, the characteristic function $\phi$ should meet all the following criteria: 
\begin{itemize}
\item \textbf{(a)} the final score benefits from strongly corresponding semantic pairs $\{v^{+}_{f}, t^{+}_{w}\}$, \ie $\phi(\mathcal{N})-\phi(\mathcal{N}\setminus\{v^{+}_{f}, t^{+}_{w}\}\cup \{[\{v^{+}_{f}, t^{+}_{w}\}]\})\textless 0$; 
\item \textbf{(b)} the final score is compromised by semantically irrelevant pairs $\{v^{-}_{f}, t^{-}_{w}\}$, \ie $\phi(\mathcal{N})-\phi(\mathcal{N}\setminus\{v^{-}_{f}, t^{-}_{w}\}\cup \{[\{v^{-}_{f}, t^{-}_{w}\}]\})\textgreater 0$; 
\item \textbf{(c)} when there are no players to cooperate, the final score is zero, \ie $\phi(\{v^{i}_{f}\}^{N_v}_{i=1})=\phi(\{t^{j}_{w}\}^{N_t}_{j=1})=\phi(\emptyset)=0$, where $\emptyset$ denotes the empty set.
\end{itemize}

Note that any function satisfying the above conditions can be used as the characteristic function $\phi$. For simplicity, we use cross-modality similarity measurement $\textrm{S}$ as $\phi$. Subsequently, we leverage Banzhaf Interaction to value possible correspondence between video frames and text words and to enhance cross-modal representation learning.

\begin{algorithm}[t]
        \caption{Multivariate Cooperative Game}
        \label{alg: code}
        \definecolor{codeblue}{rgb}{0.25,0.5,0.5}
        \definecolor{codekw}{rgb}{0.85, 0.18, 0.50}
        \begin{algorithmic}[1]
        \footnotesize
        \REQUIRE ${\bm{V}}_{f}^{s} = \{\hat{v}^{i}_{f}\}_{i=1}^{N_v}$ (original video representation), ${\bm{T}}_{w}^{s}=\{\hat{t}^{j}_{w}\}_{j=1}^{N_t}$ (original text representation)
        \STATE \# Entity level
        \STATE {Calculate ${\bm{V}}_{f}^{c}$, then combine ${\bm{V}}_{f}^{s}$ and ${\bm{V}}_{f}^{c}$ to obtain ${\bm{V}}_{f}$}
        \STATE {Calculate ${\bm{T}}_{w}^{c}$, then combine ${\bm{T}}_{w}^{s}$ and ${\bm{T}}_{w}^{c}$ to obtain ${\bm{T}}_{w}$}
        \STATE Use {{${\bm{V}}_{f}$}} and {{${\bm{T}}_{w}$}} to predict Banzhaf Interaction and calculate Banzhaf Interaction loss $\mathcal{L}_{I}^{e}$
        \STATE
        \STATE \# Action level
        \STATE Obtain coalitions ${\bm{C}}_{v}$ from ${\bm{V}}_{f}^{s}$ using DPC-KNN 
        \STATE Obtain coalitions ${\bm{C}}_{t}$ from ${\bm{T}}_{w}^{s}$ using DPC-KNN
        \STATE Utilize ${\bm{C}}_{v}$ as queries, with ${\bm{V}}_{f}^{s}$ serving as keys and values, to obtain ${\bm{V}}_{a}^{s}$ through Attention
        \STATE Utilize ${\bm{C}}_{t}$ as queries, with ${\bm{T}}_{w}^{s}$ serving as keys and values, to obtain ${\bm{T}}_{a}^{s}$ through Attention
        \STATE {Calculate ${\bm{V}}_{a}^{c}$, then combine ${\bm{V}}_{a}^{s}$ and ${\bm{V}}_{a}^{c}$ to obtain ${\bm{V}}_{a}$}
        \STATE {Calculate ${\bm{T}}_{a}^{c}$, then combine ${\bm{T}}_{a}^{s}$ and ${\bm{T}}_{a}^{c}$ to obtain ${\bm{T}}_{a}$}
        \STATE Use {{${\bm{V}}_{a}$}} and {{${\bm{T}}_{a}$}} to predict Banzhaf Interaction and calculate Banzhaf Interaction loss $\mathcal{L}_{I}^{a}$
        \STATE
        \STATE \# Event level
        \STATE Obtain coalitions ${\bm{C}}_{v}{'}$ from ${\bm{V}}_{a}^{s}$ using DPC-KNN 
        \STATE Obtain coalitions ${\bm{C}}_{t}{'}$ from ${\bm{T}}_{a}^{s}$ using DPC-KNN
        \STATE Utilize ${\bm{C}}_{v}{'}$ as queries, with ${\bm{V}}_{a}^{s}$ serving as keys and values, to obtain ${\bm{V}}_{v}^{s}$ through Attention
        \STATE Utilize ${\bm{C}}_{t}{'}$ as queries, with ${\bm{T}}_{a}^{s}$ serving as keys and values, to obtain ${\bm{T}}_{v}^{s}$ through Attention
        \STATE {Calculate ${\bm{V}}_{v}^{c}$, then combine ${\bm{V}}_{v}^{s}$ and ${\bm{V}}_{v}^{c}$ to obtain ${\bm{V}}_{v}$}
        \STATE {Calculate ${\bm{T}}_{v}^{c}$, then combine ${\bm{T}}_{v}^{s}$ and ${\bm{T}}_{v}^{c}$ to obtain ${\bm{T}}_{v}$}
        \STATE Use {{${\bm{V}}_{v}$}} and {{${\bm{T}}_{v}$}} to predict Banzhaf Interaction and calculate Banzhaf Interaction loss $\mathcal{L}_{I}^{v}$
        \RETURN $\mathcal{L}_{I}^{e}$, $\mathcal{L}_{I}^{a}$, $\mathcal{L}_{I}^{v}$
        \end{algorithmic}
    \end{algorithm}
    
\subsubsection{Multivariate Cooperative Game Modeling}
In the following, we first introduce the basic two-player interaction between video frames and word tokens. Subsequently, we expand the two-player interaction to the multivariate interaction using the token merge module. To make our module clear, we provide pseudo-code in Alg.~\ref{alg: code}.

For a coalition $\{v^{i}_{f}, t^{j}_{w}\}$, referring to Eq.~\ref{BI}, we can calculate the Banzhaf Interaction $\mathcal{I}([\{v^{i}_{f}, t^{j}_{w}\}])$. Due to the disparity in semantic similarity and interaction index, we design a prediction header to predict the fine-grained relationship $\mathcal{R}_{i,j}$ between the $i_{th}$ video frame and the $j_{th}$ text word. The prediction header consists of a convolutional layer for encoding, a self-attention module for capturing global interaction, and a convolutional layer for decoding. In Table~\ref{tab: header}, we provide the experiment results of the prediction header with different structures.

Subsequently, we optimize the Kullback-Leibler (KL) divergence~\cite{kullback1997information} between the $\mathcal{I}([\{v^{i}_{f}, t^{j}_{w}\}])$ and $\mathcal{R}_{i,j}$. Concretely, we define the probability distribution of the video-to-text task and the text-to-video task as:
\begin{equation}
\begin{aligned}
\mathcal{D}_{v2t}^{\mathcal{I}}=[p_{i,1}^{\mathcal{I}},&p_{i,2}^{\mathcal{I}},...,p_{i,N_t}^{\mathcal{I}}],\
\mathcal{D}_{t2v}^{\mathcal{I}}=[\hat p_{1,j}^{\mathcal{I}},\hat p_{2,j}^{\mathcal{I}},...,\hat p_{N_v,j}^{\mathcal{I}}], \\
p_{i,j}^{\mathcal{I}}&=\frac{\textrm{exp}{\Bigl(}\mathcal{I}([\{v^{i}_{f}, t^{j}_{w}\}]){\Bigr)}}{\sum_{k=1}^{N_t} \textrm{exp}{\Bigl(}\mathcal{I}([\{v^{i}_{f}, t^{k}_{w}\}]){\Bigr)}}, \\
\hat p_{i,j}^{\mathcal{I}}&=\frac{\textrm{exp}{\Bigl(}\mathcal{I}([\{v^{i}_{f}, t^{j}_{w}\}]){\Bigr)}}{\sum_{k=1}^{N_v} \textrm{exp}{\Bigl(}\mathcal{I}([\{v^{k}_{f}, t^{j}_{w}\}]){\Bigr)}}.
\end{aligned}
\label{eq: Dis}
\end{equation}
Similarly, the probability distribution $\mathcal{D}_{v2t}^{\mathcal{R}}$ and $\mathcal{D}_{t2v}^{\mathcal{R}}$ are calculated in the same way using $\mathcal{R}^{i,j}$:
\begin{equation}
\begin{aligned}
\mathcal{D}_{v2t}^{\mathcal{R}}=[p_{i,1}^{\mathcal{R}},p_{i,2}^{\mathcal{R}},&...,p_{i,N_t}^{\mathcal{R}}], \ \mathcal{D}_{t2v}^{\mathcal{R}}=[\hat p_{1,j}^{\mathcal{R}},\hat p_{2,j}^{\mathcal{R}},...,\hat p_{N_v,j}^{\mathcal{R}}],\\
p_{i,j}^{\mathcal{R}}&=\frac{\textrm{exp}(\mathcal{R}_{i,j})}{\sum_{k=1}^{N_t} \textrm{exp}(\mathcal{R}_{i,k})}, \\ \hat p_{i,j}^{\mathcal{R}}&=\frac{\textrm{exp}(\mathcal{R}_{i,j})}{\sum_{k=1}^{N_v} \textrm{exp}(\mathcal{R}_{k,j})}.
\end{aligned}
\end{equation}
Finally, the Banzhaf Interaction loss $\mathcal{L}_{I}$ is defined as:
\begin{equation}
\mathcal{L}_{I}=\mathbb{E}_{\bm{v},\bm{t}} \Bigl[\textrm{KL}(\mathcal{D}_{v2t}^{\mathcal{R}}\|\mathcal{D}_{v2t}^{\mathcal{I}})+\textrm{KL}(\mathcal{D}_{t2v}^{\mathcal{R}}\|\mathcal{D}_{t2v}^{\mathcal{I}})\Bigr].
\end{equation}
The Banzhaf Interaction loss $\mathcal{L}_{I}$ brings the probability distributions of the output $\mathcal{R}$ of the prediction header and Banzhaf Interaction $\mathcal{I}$ close together to establish fine-grained semantic alignment between video frames and text words. In particular, it can be directly removed during inference, rendering an efficient and semantics-sensitive model.

\begin{figure}[tbp]
\centering
    \includegraphics[width=1.\linewidth]{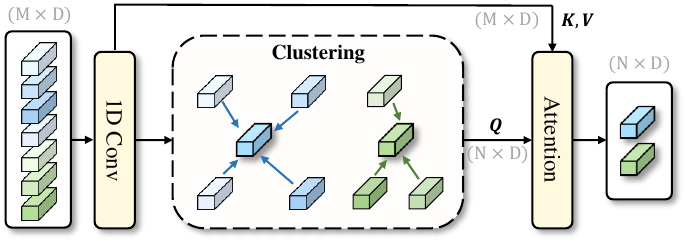}
    \vspace{-0.2in}
    \caption{\textbf{The token merge module.} ``1D-Conv'' denotes the one-dimensional convolutional layer. $M$ input tokens with $D$ channels are first clustered into $N$ clusters. Subsequently, we feed the merged tokens as queries $Q$ and the original tokens as keys $K$ and values $V$ into an attention module.}
    \label{method: fig3}
    \vspace{-0.1in}
\end{figure}

For multivariate interaction, an intuitive method is to compute Banzhaf Interaction on any candidate set of visual frames and text words directly. However, the number of candidate sets is too large, \ie $2^{N_v+N_t}$. To reduce the number of candidate sets, we cluster the original visual (textual) tokens and compute the Banzhaf Interaction between the merged tokens. By stacking token merge modules, we get cross-modal interaction efficiently at different semantic levels, \ie entity-level interactions on the frames and words, action-level interactions on the clips and phrases, and event-level interactions on the segments and paragraphs. Fig.~\ref{method: fig3} illustrates the framework of the token merge module.

Specifically, we utilize DPC-KNN~\cite{du2016study}, a k-nearest neighbor-based density peaks clustering algorithm, to cluster the visual (textual) tokens. Starting with the frame-level tokens $\bm{V}_{f}^{s}=\{\hat{v}^{i}_{f}\}^{N_v}_{i=1}$, we first use a one-dimensional convolutional layer to enhance the temporal information between tokens. Subsequently, we compute the local density $\rho_i$ of each token $\hat{v}^{i}_{f}$ according to its $K$-nearest neighbors:
\begin{equation}
{
\rho_i=\textrm{exp}{\Bigl(}-\frac{1}{K}\sum_{\hat{v}^{k}_{f}\in \textrm{KNN}(\hat{v}^{i}_{f})}\Vert \hat{v}^{k}_{f}-\hat{v}^{i}_{f} \Vert^2 {\Bigr)},
}
\end{equation}
where $\textrm{KNN}(\hat{v}^{i}_{f})$ is the $K$-nearest neighbors of $\hat{v}^{i}_{f}$. 
After that, we compute the distance index {$\xi_i$} of each token $\hat{v}^{i}_{f}$:
  \begin{equation}
    {
    {\xi_i}=
    \begin{cases}
    \underset{j:\rho_j>\rho_i}{\textrm{min}} \Vert \hat{v}^{j}_{f}-\hat{v}^{i}_{f} \Vert^2, & \text{if $\exists j$ s.t. $\rho_j>\rho_i$.}\\
    \ \ \underset{j}{\textrm{max}} \ \ \Vert \hat{v}^{j}_{f}-\hat{v}^{i}_{f} \Vert^2, & \text{otherwise.}
    \end{cases}
    }
  \end{equation}
Intuitively, $\rho$ denotes the local density of tokens, and {$\xi$} represents the distance from other high-density tokens. 

\begin{figure*}
\centering
\includegraphics[width=1.\linewidth]{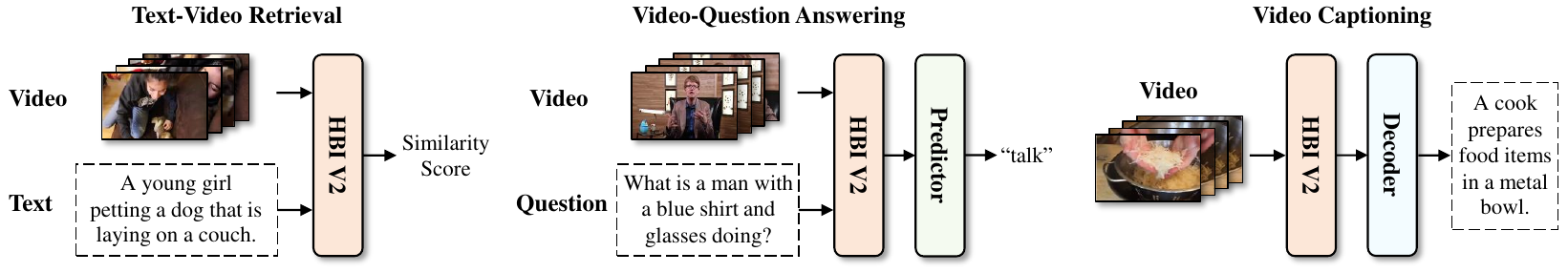}
\vspace{-0.21in}
\caption{\textbf{The task-specific prediction heads.} To enhance the versatility of HBI V2 across various downstream tasks, we expand our original structure into a flexible encoder-decoder framework comprising an encoder and a task-specific decoder.}
\label{method: fig4}
\vspace{-0.12in}
\end{figure*}

We consider those tokens with relatively high {$\rho_i \times \xi_i$} as cluster centers and then assign other tokens to the nearest cluster center according to the Euclidean distances. Inspired by~\cite{li2022dynamic,rao2021dynamicvit}, we use the weighted average tokens of each cluster to represent the corresponding cluster, where the weight $W=\textrm{MLP}_{w}(\bm{V}_{f}^{s})$. Subsequently, we feed the weighted average tokens as queries $Q$ and the original tokens as keys $K$ and values $V$ into an attention module. 

To make the token with high weight have a high contribution to the output, we incorporate the weight $W$ into the attention module as follows:
\begin{equation}
{\textrm{Attention}(Q,K,V,W)=\textrm{Softmax}\Bigl(\frac{QK^\top}{\sqrt{d_k}}+W \Bigr)V,}
\end{equation}
where $d_k$ is the channel number of the queries and is used as the scaling factor. We treat the output of the attention module as features at a higher semantic level than the entity level, that is, the action-level visual tokens. Similarly, we merge the action-level tokens again to get the event-level tokens. The action-level textual tokens and event-level textual tokens are calculated in the same way.

\subsection{Task-Specific Prediction Heads}
As shown in Fig.~\ref{method: fig4}, to enhance the versatility of HBI V2 across various downstream tasks, including VideoQA and video captioning, we expand our original HBI structure into a flexible encoder-decoder framework. This framework consists of an encoder and a task-specific decoder, allowing for efficient adaptation to diverse tasks.

In text-to-video retrieval, given a text query alongside a gallery of videos, the objective is to rank all videos so that the video corresponding to the text query is ranked as high as possible. Similarly, in video-to-text retrieval, the goal is to rank all text candidates based on the video query. In our HBI V2 framework, we can directly rank candidates by leveraging the similarity scores between video and text, eliminating the necessity for an additional prediction head.

For video-question answering, due to the established video-text fine-grained alignment from the hierarchical Banzhaf Interaction module, we can adopt a simplified answer prediction head, without the need for sophisticated multi-modal fusion/reasoning stages like many previous VideoQA models. For the question-answering prediction head, we concatenate the video representation and text representation followed by an MLP to predict the answer.

For video captioning, we utilize a single-layer transformer decoder as the generator of the caption. The decoder sequentially produces hidden features. Subsequently, we employ a linear projection layer to map these hidden features to the vocabulary dictionary. Notably, given the absence of text input in the video captioning task, we rely on single-modal representations instead of reconstructed representations during inference. However, it is worth noting that the reconstructed representations are still utilized for video-language alignment purposes during training.

\begin{table*}[t]
\caption{\textbf{Comparisons to current text-video retrieval methods on the MSRVTT, ActivityNet Captions, and DiDeMo datasets.} ``$\uparrow$'' denotes that higher is better. ``$\downarrow$'' denotes that lower is better. All results do not use inverted softmax~\cite{bogolin2022cross}.}
\label{tab: Comparisons to State-of-the-arts}
\subfloat[\normalsize{Text-to-video retrieval and video-to-text retrieval performance on the MSRVTT dataset.}]
{
\resizebox{1.\linewidth}{!}{
\begin{tabular}{l|cccccc|cccccc}
\toprule[1.25pt]
\multirow{2}{*}{\textbf{Methods}} &\multicolumn{6}{c|}{\textbf{Text-to-Video}} &\multicolumn{6}{c}{\textbf{Video-to-Text}}\\
\cmidrule(rl){2-7}\cmidrule(rl){8-13}
  & R@1 $\uparrow$ & R@5 $\uparrow$ & R@10 $\uparrow$ & Rsum $\uparrow$ & MdR $\downarrow$ & MnR $\downarrow$ & R@1 $\uparrow$ & R@5 $\uparrow$ & R@10 $\uparrow$ & Rsum $\uparrow$ & MdR $\downarrow$ & MnR $\downarrow$ \\ \midrule
MMT~{\cite{gabeur2020multi}}~\pub{ECCV 2020} & {26.6} & 57.1 & 69.6 & 153.3 & 4.0 & 24.0 & 27.0 & 57.5 & 69.7 & 154.2 & 3.7 & 21.3 \\
T2VLAD~{\cite{wang2021t2vlad}}~\pub{CVPR 2021} & {29.5} & 59.0 & 70.1 & 158.6 & 4.0 & -  & 31.8 &60.0 &71.1 & 162.9 &\Scnd{3.0} & -\\
TT-CE~{\cite{croitoru2021teachtext}}~\pub{ICCV 2021} & {29.6} & 61.6 & 74.2 & 165.4 & \Scnd{3.0} & - & 32.1 & 62.7 & 75.0 & 169.8 &\Scnd{3.0} & -\\
Support-Set~{\cite{patrick2021support}}~\pub{ICLR 2021}  & {30.1} & 58.5 & 69.3 & 157.9 & \Scnd{3.0} & - &28.5 &58.6 &71.6 & 158.7 &\Scnd{3.0} & - \\
CLIP4Clip~{\cite{luo2022clip4clip}}~\pub{Neurocomputing 2022} & {44.5} & 71.4 & 81.6 & 197.5 &\Frst{2.0} & 15.3 & 42.7 & 70.9& 80.6 & 194.2 & \Frst{2.0} & 11.6\\
EMCL~{\cite{jin2022expectation}}~\pub{NeurIPS 2022}  & {46.8} & 73.1 &\Scnd{83.1} & 203.0 &\Frst{2.0} & 12.8 & 
46.5 & 73.5 & 83.5 & 203.5 & \Frst{2.0} & 8.8 \\
X-Pool~{\cite{gorti2022x}}~\pub{CVPR 2022}  & {46.9} & 72.8 & 82.2 & 201.9 &\Frst{2.0} & 14.3 & 44.4 & 73.3 & 84.0 & 201.7 & \Frst{2.0} & 9.0 \\
TS2-Net~{\cite{liu2022ts2}}~\pub{ECCV 2022}  & {47.0} & \Scnd{74.5} & \Frst{83.8} &{205.3} & \Frst{2.0} & \Scnd{13.0} & 45.3 & 74.1 & 83.7 & 203.1 & \Frst{2.0} & 9.2 \\
UATVR~{\cite{fang2023uatvr}}~\pub{ICCV 2023}   & {47.5} & 73.9 & 83.5 & 204.9 &\Frst{2.0} & 12.3 & 
46.9 & 73.8 & 83.8 & 204.5 & \Frst{2.0} & 8.6 \\
Prompt Switch~{\cite{deng2023prompt}}~\pub{ICCV 2023}  & {47.8} & 73.9 & 82.2 & 203.9 &\Frst{2.0} & 14.1 & 
46.0 & \textbf{74.3} & \textbf{84.8} & 205.1 & \Frst{2.0} & \textbf{8.5} \\
\midrule
HBI & 48.6 & 74.6 & 83.4 & 206.6 & \textbf{2.0} & \textbf{12.0} & 46.8 & \textbf{74.3} & 84.3 & 205.4 & \textbf{2.0} & 8.9 \\
\rowcolor{aliceblue!60} \textbf{HBI V2} & \textbf{49.4} & \textbf{74.6} & \textbf{83.8} & \textbf{207.8} & \textbf{2.0} &{12.2} & \textbf{47.7} & 74.2 &{84.4} & \textbf{206.3} &\textbf{2.0} &\Scnd{8.9} \\
\bottomrule[1.25pt]
\end{tabular}
}
}\\
\subfloat[\normalsize{Text-to-video retrieval performance on the ActivityNet Captions and DiDeMo datasets.}]
{
\centering
\resizebox{1.\linewidth}{!}
{
\begin{tabular}{l|cccccc|cccccc}
\toprule[1.25pt]
\multirow{2}{*}{\textbf{Method}} &  \multicolumn{6}{c|}{\textbf{ActivityNet Captions}}        & \multicolumn{6}{c}{\textbf{DiDeMo}}  \\ 
\cmidrule(rl){2-7}\cmidrule(rl){8-13}
  & R@1 $\uparrow$ & R@5 $\uparrow$ & R@10 $\uparrow$ & Rsum $\uparrow$ & MdR $\downarrow$ & MnR $\downarrow$
& R@1 $\uparrow$ & R@5 $\uparrow$ & R@10 $\uparrow$ & Rsum $\uparrow$ & MdR $\downarrow$ & MnR $\downarrow$    \\ \midrule
ClipBERT~\cite{lei2021less}~\pub{CVPR 2021} & 21.3 & 49.0 & 63.5 & 133.8 & 6.0 & - & 20.4 & 48.0 & 60.8 & 129.2 & 6.0 & - \\
Frozen~\cite{bain2021frozen}~\pub{ICCV 2021} & - & - & - & - & - & - & 34.6 & 65.0 & 74.7 & 174.3 & 3.0 & - \\
CLIP4Clip~{\cite{luo2022clip4clip}}~\pub{Neurocomputing 2022}  & 40.5 & 72.4 & 83.6 & 196.5 & \Frst{2.0} & 7.5 & 42.8 & 68.5 & 79.2 & 190.5 & \Frst{2.0} & 18.9 \\
TS2-Net~{\cite{liu2022ts2}}~\pub{ECCV 2022}  & 41.0 & {73.6} & 84.5 & 199.1 & \Frst{2.0} & 8.4 & 41.8 & 71.6 & 82.0 & 195.4 & \Frst{2.0} & 14.8 \\
UATVR~{\cite{fang2023uatvr}}~\pub{ICCV 2023} & - & - & - & - & - & - & 43.1 & 71.8 & 82.3 & 197.2 & \textbf{2.0} & 15.1 \\
EMCL~{\cite{jin2022expectation}}~\pub{NeurIPS 2022}  & 41.2 & 72.7 & 83.6 & 197.5 & \textbf{2.0} & 8.6 & 45.3 & 74.2 & 82.3 & 201.8 & \textbf{2.0} & 12.3 \\
CenterCLIP~\cite{zhao2022centerclip}~\pub{SIGIR 2022} & 43.9 & 74.6 & 85.8 & 204.3 & \textbf{2.0} & 6.7 & - & - & - & - & - & - \\
\midrule
HBI & 42.2 & 73.0 & 84.6 & 199.8 & \textbf{2.0} & 6.6 & 46.9 & \textbf{74.9} & 82.7 & 204.5 & \textbf{2.0} & \textbf{12.1} \\ 
\rowcolor{aliceblue!60} \textbf{HBI V2} & \textbf{45.5} & \textbf{75.3} & \textbf{86.0} & \textbf{206.8} &\textbf{2.0} & \textbf{6.5} & \textbf{47.4} & {74.6} & \textbf{84.5} & \textbf{206.5} &\textbf{2.0} & {12.7} \\ 
\bottomrule[1.25pt]
\end{tabular}
}
}
\vspace{-0.2in}
\end{table*}

\subsection{Training Objective}
To achieve fine-grained semantic alignment, we first employ cross-modal contrastive loss. Given video representation $\bm{V}_{f}=\{v^{i}_{f}\}^{N_v}_{i=1}$ and text representation $\bm{T}_{w}=\{t^{j}_{w}\}^{N_t}_{j=1}$, the alignment matrix is defined as: $A = [a_{ij}]^{N_{v}\times N_{t}}$, where $a_{ij}=\frac{(v^{i}_{f})^\top t^{j}_{w}}{\Vert v^{i}_{f}\Vert \Vert t^{j}_{w}\Vert}$ represents the alignment score between the $i_{th}$ video frame and the $j_{th}$ text word. For the $i_{th}$ video frame, we calculate its maximum alignment score as $\underset{j}{\textrm{max}}\ a_{ij}$. Subsequently, we use the weighted average maximum alignment score over all video frames as the video-to-text similarity. Similarly, we can obtain the text-to-video similarity. The total similarity score is defined as: 
\begin{equation}
\textrm{S}_{\bm{v},\bm{t}}=\frac{1}{2}{\Bigl(}\underbrace{\sum_{i=1}^{N_v} \omega_v^i\ \underset{j}{\textrm{max}}\ a_{ij}}_{\textrm{video-to-text similarity}}+\underbrace{\sum_{j=1}^{N_t} \omega_t^j\ \underset{i}{\textrm{max}}\ a_{ij}}_{\textrm{text-to-video similarity}}{\Bigr)},
\end{equation}
where $[\omega_v^0,\omega_v^1,...,\omega_v^{N_v}]=\textrm{Softmax}{\Bigl(}\textrm{MLP}_v(\bm{V}_{f}){\Bigr)}$ and $[\omega_t^0,\omega_t^1,...,\omega_t^{N_t}]=\textrm{Softmax}{\Bigl(}\textrm{MLP}_t(\bm{T}_{w}){\Bigr)}$ are the weights of the video frames and text words, respectively. Then the cross-modal contrastive loss~\cite{van2018representation} is formulated as:
\begin{equation}
\begin{aligned}
\mathcal{L}_{C}=-\frac{1}{2}{\Bigl[}\frac{1}{B}\sum_{k=1}^{B}\log\frac{\exp(\textrm{S}_{\bm{v}_k,\bm{t}_k}/\tau)}{\sum_{l}^{B}\exp(\textrm{S}_{\bm{v}_k,\bm{t}_l}/\tau)} +\\  \frac{1}{B}\sum_{k=1}^{B}\log\frac{\exp(\textrm{S}_{\bm{v}_k,\bm{t}_k}/\tau)}{\sum_{l}^{B}\exp(\textrm{S}_{\bm{v}_l,\bm{t}_k}/\tau)}{\Bigr]},
\end{aligned}
\end{equation}
where $B$ is the batch size and $\tau$ is the temperature hyper-parameter. This loss function maximizes the similarity of positive pairs and minimizes the similarity of negative pairs.

Combining the cross-modal contrastive loss $\mathcal{L}_{C}$ and Banzhaf Interaction loss $\mathcal{L}_{I}$, the full objective of semantic alignment is formulated as $\mathcal{L}=\mathcal{L}_{C}+\alpha \mathcal{L}_{I}$, where $\alpha$ is the trade-off hyper-parameter. We train the network at three semantic levels, which are shown as follows:
\begin{equation}
\mathcal{L}^e\!=\!\mathcal{L}_{C}^e\!+\!\alpha \mathcal{L}_{I}^e,\quad
\mathcal{L}^a\!=\!\mathcal{L}_{C}^a\!+\!\alpha \mathcal{L}_{I}^a,\quad
\mathcal{L}^v\!=\!\mathcal{L}_{C}^v\!+\!\alpha \mathcal{L}_{I}^v,
\label{eq: loss0}
\end{equation}
where $\mathcal{L}^e$, $\mathcal{L}^a$, and $\mathcal{L}^v$ represent the semantic alignment loss at the entity level, action level, and event level, respectively.

To further improve the generalization ability, we optimize the additional KL divergence between the distribution among different semantic levels. We find that the entity-level similarity $\textrm{S}^e_{\bm{v},\bm{t}}$ converges first in the training process, so we distill the entity-level similarity to the other two semantic levels. Starting with entity-level similarity $\textrm{S}^e_{\bm{v},\bm{t}}$ distilling to action-level similarity $\textrm{S}^a_{\bm{v},\bm{t}}$, we first calculate the distribution $\mathcal{D}_{v2t}^{e}$ and $\mathcal{D}_{t2v}^{e}$ by replacing $\mathcal{I}([\{v, t\}])$ with $\textrm{S}^e_{\bm{v},\bm{t}}$ in Eq.~\ref{eq: Dis}. The distribution $\mathcal{D}_{v2t}^{a}$ and $\mathcal{D}_{t2v}^{a}$ are calculated using $\textrm{S}^a_{\bm{v},\bm{t}}$ in the same way. The $\mathcal{L}_{D}^{e2a}$ loss is defined as:
\begin{equation}
\mathcal{L}_{D}^{e2a}=\mathbb{E}_{\bm{v},\bm{t}} {\Bigl[}\textrm{KL}(\mathcal{D}_{v2t}^{a}\|\mathcal{D}_{v2t}^{e})+\textrm{KL}(\mathcal{D}_{t2v}^{a}\|\mathcal{D}_{t2v}^{e}){\Bigr]}.
\label{eq: loss1}
\end{equation}
The $\mathcal{L}_{D}^{e2v}$ loss from entity-level similarity to event-level similarity is calculated in the same way.

Furthermore, depending on the specific task, we may require the inclusion of task-specific loss. Particularly, in the case of text-video retrieval, there is no necessity to introduce a task-specific loss. However, for tasks such as VideoQA and video captioning, an additional cross-entropy loss $\mathcal{L}_{task}$ needs to be incorporated. The overall loss is the combination of semantically alignment losses, self-distillation losses, and task-specific loss, which is defined as:
\begin{equation}
\mathcal{L}_{total}=\underbrace{\mathcal{L}^{e}+\mathcal{L}^{a}+\mathcal{L}^{v}}_{\textrm{deep supervision}}+\beta \underbrace{(\mathcal{L}_{D}^{e2a}+\mathcal{L}_{D}^{e2v})}_{\textrm{self-distillation}}+\lambda\mathcal{L}_{task},
\end{equation}
where $\beta$ is the trade-off hyper-parameter. For VideoQA and video captioning, we apply $\lambda$ to balance the feature alignment penalty and the task-specific penalty.

\begin{figure*}[ht]
    \centering
    \includegraphics[width=1.\textwidth]{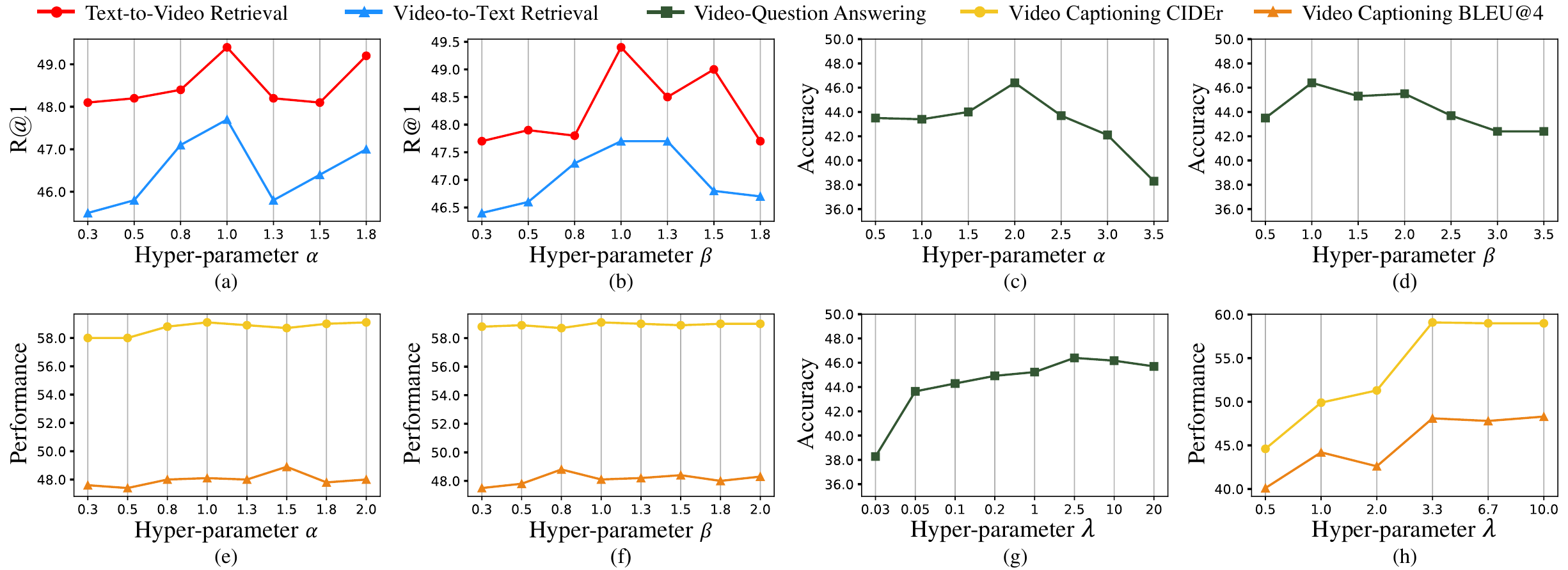}
    \vspace{-0.23in}
    \caption{{\textbf{Parameter sensitivity.} $\alpha$, $\beta$, and $\lambda$ are the trade-off hyper-parameters in Eq.~\ref{eq: loss0} and Eq.~\ref{eq: loss1}. (a) and (b) are the effects of hyper-parameters $\alpha$ and $\beta$ on the MSRVTT dataset for the text-video retrieval task. (c) and (d) are the effects of hyper-parameters $\alpha$ and $\beta$ on the MSRVTT-QA dataset for the video-question answering task. (e) and (f) are the effects of hyper-parameters $\alpha$ and $\beta$ on the MSRVTT dataset for the video captioning task. (g) and (h) are the effects of hyper-parameter $\lambda$ on the MSRVTT-QA and MSRVTT datasets for the video-question answering and video captioning tasks.}}
    \label{exp: fig1}
    \vspace{-0.1in}
\end{figure*}

\section{Experiments}
\subsection{Experimental Settings}
\myparagraph{Text-Video Retrieval Datasets.} 
\textbf{MSRVTT}~\cite{xu2016msr} contains 10K YouTube videos, each video with 20 text descriptions. We follow the training protocol in \cite{liu2019use,gabeur2020multi,miech2019howto100m} and evaluate text-to-video and video-to-text retrieval tasks on the 1K-A testing split with 1K video or text candidates defined by \cite{yu2018a}.
\textbf{ActivityNet Captions}~\cite{krishna2017dense} consists of densely annotated temporal segments of 20K YouTube videos. Following~\cite{gabeur2020multi,patrick2021support,wang2021t2vlad}, we concatenate descriptions of segments in a video to construct ``video-paragraph'' for retrieval. We use the 10K training split to finetune the model and report the performance on the 5K ``val1'' split.
\textbf{DiDeMo}~\cite{anne2017localizing} contains 10K videos annotated with 40K text descriptions. We concatenate descriptions of segments in a video to construct a ``video paragraph'' for retrieval. We follow the training and evaluation protocol in CLIP4Clip~\cite{luo2022clip4clip}.

\myparagraph{Video-Question Answering Datasets.}
\textbf{MSRVTT-QA}~\cite{xu2017video} comprises 10K videos and 243K question-answer pairs. The video has a longer duration of 10-30 seconds.
\textbf{MSVD-QA}~\cite{xu2017video} comprises 1,970 short clips~(10 seconds on average) and 50,505 question-answer pairs. 
\textbf{ActivityNet-QA}~\cite{yu2019activitynet} consists of 58,000 QA pairs on 5,800 complex long web videos. The average video length of ActivityNet-QA is 180 seconds, longer than MSRVTT-QA and MSVD-QA.

\myparagraph{Video Captioning Dataset.}
The \textbf{MSRVTT-Caption}~\cite{xu2016msr} in the video captioning task is the same as the MSRVTT dataset in the text-video retrieval task.

\myparagraph{Evaluation Metrics.} We choose Recall at rank K (R@K), Median Rank (MdR), and mean rank (MnR) to evaluate the retrieval performance. We select the answer accuracy to evaluate the video-question answering performance. We apply four metrics for the video caption task, including BLEU-4~\cite{papineni2002bleu}, ROUGE-L~\cite{luo2020univl}, METEOR~\cite{banerjee2005meteor}, and CIDEr~\cite{vedantam2015cider}.

\myparagraph{Implementation Details.}
We utilize the CLIP~\cite{radford2021learning} as the pre-trained model. We use the Adam optimizer~\cite{kingma2014adam} and set the temperature $\tau$ to 0.01. Since the calculation of the exact Banzhaf Interaction is an NP-hard problem~\cite{matsui2001np}, existing methods mainly use sampling-based methods~\cite{bachrach2010approximating} to obtain unbiased estimates. To speed up the computation of Banzhaf Interaction for many data instances, we pre-train a tiny model to learn a mapping from a set of input features to a result using MSE loss. The tiny model consists of 2 CNN layers and a self-attention layer. The input is the similarity matrix of video frames and text tokens, and the output is the estimation of Banzhaf Interaction. For text-video retrieval, the frame length and caption length are 12 and 32 for the MSRVTT dataset. We set the frame length and caption length to 64 and 64 for the ActivityNet Captions dataset. We set the frame length and caption length to 32 and 64 for the DiDeMo dataset. The initial learning rate is 1e-7 for text encoder and video encoder and 1e-3 for other modules. The network is optimized with a batch size of 128 in 5 epochs. For video-question answering, we use the target vocabulary and train a fully connected layer on top of the final language features to classify the answer. The initial learning rate is 1e-7 for text encoder and video encoder and 1e-3 for other modules. We set the task balance factor $\lambda$ to 2.5. The network is optimized with a batch size of 32 in 5 epochs. For video captioning, we initialize the learning rate at 5e-4 and freeze the CLIP encoder. The network is optimized with a batch size of 64 in 50 epochs.

\begin{table*}[t]
\begin{minipage}[c]{0.48\textwidth}
\centering
{
\caption{\textbf{Comparisons to VideoQA methods on the MSRVTT-QA, MSVD-QA, and ActivityNet-QA datasets.} We gray out pretrained models to ensure a fair comparison.}\label{tab: videoqa}
    \centering
    \vspace{-.25em}
    \setlength{\tabcolsep}{1.5mm}
    \begin{tabular}{lccc}
        \toprule[1.25pt]
        \multirow{2}{*}{\textbf{Methods}} & {\textbf{MSRVTT-QA}} & {\textbf{MSVD-QA}} & {\textbf{ActivityNet-QA}} \\
        \cmidrule(rl){2-4}
        & Accuracy $\uparrow$ & Accuracy $\uparrow$ & Accuracy $\uparrow$ \\ \midrule
        \color{gray}{Co-Tok~\cite{piergiovanni2022video}} & \color{gray}{45.7} & \color{gray}{48.6} & \color{gray}{-}\\
        \color{gray}{All-in-One~\cite{wang2023all}} & \color{gray}{46.8} & \color{gray}{48.3} & \color{gray}{-}  \\
        \color{gray}{FrozenBiLM~\cite{yang2022zero}} & \color{gray}{47.0} & \color{gray}{54.8} & \color{gray}{43.2}\\
        \midrule
        MGIN~\cite{wang2023multi} & 38.2 & 39.7 & - \\
        IGV~\cite{li2022invariant} & 38.3 & 40.8 & -    \\
        CLIP-QA~\cite{radford2021learning} & 39.0 & 38.5 & -\\
        VQA-T~\cite{yang2022learning} & 39.6 & 41.2 & 36.8 \\
        DRV~\cite{liu2023video} & 40.0 & 42.2 & - \\
        CLIP4Clip~\cite{luo2022clip4clip} & 40.9 & 39.3 & - \\
        SViTT~\cite{li2023svitt} & 43.2 & - & 43.0 \\
        TG-VQA~\cite{li2023tg} & 46.3 & 52.5 & 48.3 \\
        \midrule
         {HBI} & {46.2} & {52.1} & {48.2} \\
        \rowcolor{aliceblue!60} \textbf{HBI V2} & \Frst{46.4} & \Frst{53.4} & \textbf{49.3}  \\
        \bottomrule[1.25pt]
    \end{tabular}
}
\end{minipage}
\hfill
\begin{minipage}[l]{0.49\textwidth}
{
\caption{\textbf{Comparisons to current video captioning methods on the MSRVTT dataset.} We gray out pretrained models to ensure a fair comparison.}
        \label{tab: video caption}
    \vspace{-.25em}
    \centering
    \setlength{\tabcolsep}{1.8mm}
    \begin{tabular}{lcccc}
        \toprule[1.25pt]
        \multirow{2}{*}{\textbf{Methods}} & \multicolumn{4}{c}{\textbf{MSRVTT}}\\
        \cmidrule(rl){2-5}
        & BLEU@4 $\uparrow$ & ROUGE $\uparrow$ & METEOR $\uparrow$ & CIDEr $\uparrow$ \\ \midrule
        \color{gray}{SwinBERT~\cite{lin2022swinbert}} & \color{gray}{45.4} & \color{gray}{30.6} & \color{gray}{64.1} & \color{gray}{55.9} \\
        \color{gray}{MV-GPT~\cite{seo2022end}} & \color{gray}{48.9} &  \color{gray}{64.0} & \color{gray}{38.7} & \color{gray}{60.0} \\
        \midrule
        MGCMP~\cite{chen2021motion} & 41.7 & 62.1 & 28.9  & 51.4     \\
        OpenBook~\cite{zhang2021open} & 42.8 & 61.7 & 29.3  & 52.9    \\
        AVSSN~\cite{perez2021attentive} & 42.8 & 61.7 & 28.8 & 46.9\\
        GSEN~\cite{luo2024global} & 42.9 & 61.7 & 28.4 & 51.0 \\ 
        AFC~\cite{shen2023accurate} & {43.1} & 62.7 & {29.8} & {56.2}\\
        VEIN~\cite{song2024emotional} & 44.1 & 62.6 & 30.0 & 55.3 \\
        SemSynAN~\cite{perez2021improving} &  44.3 & 62.5 & 28.8 & 50.1\\
        EMCL~\cite{jin2022expectation} & 45.3 & 63.2 & 30.2  & 54.6 \\
        {ClipCaption}~\cite{tang2021clip4caption} & {46.1} & {63.7} & {30.7} & {57.7}\\
        \midrule
         {HBI} & {45.2} & {63.0} & {30.1} & {55.3} \\
        \rowcolor{aliceblue!60} \textbf{HBI V2} & \Frst{48.1} & \Frst{64.9} & \textbf{31.4} & \Frst{59.1} \\
        \bottomrule[1.25pt]
    \end{tabular}
}
\end{minipage}
\end{table*}

\begin{table*}[t]
\begin{minipage}[l]{0.65\textwidth}
    {
    \caption{\textbf{Ablation study about the importance of each part of our method on the MSRVTT dataset.} ``QA'' denotes the answer accuracy on MSRVTT-QA.}
    \label{tab: ablation}
    \vspace{-.25em}
    \setlength{\tabcolsep}{1.2mm}
    \begin{tabular}{ccccccccc}
    \toprule[1.25pt]
    {$\mathcal{L}_{I}$ Banzhaf} & {Deep} & {$\mathcal{L}_{D}$ Self} & Representation & \multicolumn{4}{c}{{Text-to-Video}} & \multirow{2}{*}{QA$\uparrow$} \\  \cmidrule{5-8}
    {Interaction} & {Supervision} & {Distillation} & Reconstruction & R@1$\uparrow$ & R@5$\uparrow$ & R@10$\uparrow$ & MnR$\downarrow$ &\\ \midrule
    & & & & 46.6 & 73.1 & 83.0 & 13.3 & 44.5\\
    {\Checkmark} & &  & & 47.4 & 74.2 & 82.8 & 12.1 & 45.8\\
    & {\Checkmark} &  & & 47.2 & 74.1 & 82.6 & 12.0 & 46.0\\
    & {\Checkmark} & {\Checkmark}  & & 47.6 & 73.8 & 83.2 & \textbf{11.9}&46.0\\
    {\Checkmark}  & {\Checkmark} & & & 48.2 & 73.0 & 83.1 & 12.0& 46.1 \\
    {\Checkmark} & {\Checkmark} & {\Checkmark} & & {48.6} & \textbf{74.6} & {83.4} & 12.0 & {46.2}\\
    \midrule
    \rowcolor{aliceblue!60} {\Checkmark} & {\Checkmark} & {\Checkmark}  & {\Checkmark}  & \textbf{49.4} & \textbf{74.6} & \textbf{83.8} & 12.2 & \textbf{46.4} \\
    \bottomrule[1.25pt]
    \end{tabular}
    }
\end{minipage}
\hfill
\begin{minipage}[c]{0.32\textwidth}
\centering
{
\tabcaption{\textbf{Effect of Prediction heads on the MSRVTT.} ``None'' denotes no header. ``SA'' denotes the attention module. ``CNN+SA'' performs the best.
}
\vspace{-0.45em}
\label{tab: header}
\centering
\setlength{\tabcolsep}{4.5pt}
\begin{tabular}{lcccc}
    \toprule[1.25pt]
    \multirow{2}{*}{Method} &  \multicolumn{4}{c}{{Text-to-Video}}  \\
    \cmidrule{2-5}
      & R@1$\uparrow$ & R@5$\uparrow$ & R@10$\uparrow$ & MnR$\downarrow$ \\ \midrule
    Baseline  & 46.6 & 73.1 & 83.0 & 13.3 \\
    \midrule
    None  & 47.7 & 73.7 & 83.2 & \textbf{12.2} \\
    MLP  & 48.0 & 74.1 & 82.6 & 12.5 \\
    CNN  & 48.5 & 73.9 & 82.6 & 12.6 \\
    MLP+SA  & 48.4 & 71.8 & 81.7 & 12.7 \\
    \rowcolor{aliceblue!60} CNN+SA   & \textbf{49.4} & \textbf{74.6} & \textbf{83.8} & \textbf{12.2} \\ \bottomrule[1.25pt]
\end{tabular}
}
\end{minipage}
\vspace{-0.1in}
\end{table*}

\subsection{Comparison with State-of-the-Art}
We conduct extensive experiments on three datasets for text-video retrieval, three datasets for video-question answering, and one dataset for video captioning. In Table~\ref{tab: Comparisons to State-of-the-arts}, we show the retrieval results of our method on MSRVTT, ActivityNet Captions, and DiDeMo datasets. Additionally, Table~\ref{tab: videoqa} showcases the video-question answering results on the MSRVTT-QA, MSVD-QA, and ActivityNet-QA datasets. Furthermore, in Table~\ref{tab: video caption}, we exhibit the video captioning results on the MSRVTT dataset. Our proposed HBI V2 consistently outperforms both the previous HBI and existing methods across all downstream tasks. These results demonstrate the effectiveness of our proposed HBI V2 framework. Additionally, We observe that HBI V2 exhibits notable advancements over HBI on the MSVD-QA and ActivityNet-QA, the improvement on the MSRVTT-QA appears less pronounced. We consider this may be because our framework focuses on video-language alignment, whereas the MSRVTT-QA dataset presents a greater challenge in video-language reasoning, resulting in comparatively marginal enhancements.

\subsection{Ablation Study}
\myparagraph{Ablation about Components.} To illustrate the importance of each part of our method including the Banzhaf Interaction, the deep supervision structure, the self-distillation, and the representation reconstruction, we conduct ablation experiments on both MSRVTT and MSRVTT-QA datasets in Table~\ref{tab: ablation}. The Banzhaf Interaction boosts the baseline with an improvement up to 0.8\% at R@1 and 1.3\% at answer accuracy. Furthermore, deep supervision and self-distillation significantly improve the generalization ability. Additionally, the representation reconstruction further improves the performance of the model. Our full model attains superior performance, surpassing the baseline by 2.8\% at R@1 for text-to-video retrieval and 1.9\% at answer accuracy for video-question answering. This shows that the four parts are beneficial for aligning videos and texts. 

The parameter $\alpha$ serves as the hyper-parameter that balances the cross-modality contrastive loss~($\mathcal{L}_{C}$) and the Banzhaf Interaction loss~($\mathcal{L}_{I}$). Meanwhile, the parameter $\beta$ acts as the hyper-parameter regulating the trade-off between the loss from deep supervision and the loss from self-distillation. $\lambda$ is the hyper-parameter used to balance the feature alignment penalty and the task-specific penalty. To determine the optimal settings for $\alpha$ and $\beta$ for the text-video retrieval task, we evaluate the scale range setting $\alpha \in [0.3, 1.8]$ as shown in Fig.~\ref{exp: fig1}(a). Our findings indicate that enhancing $\alpha$ from 0.5 to 0.8 leads to an improvement in R@1 from 48.2\% to 48.4\%, with saturation observed at $\alpha = 1.0$ for text-to-video retrieval. Consequently, we adopt $\alpha = 1.0$ to achieve optimal performance. In Fig.~\ref{exp: fig1}(b), we demonstrate the impact of the hyper-parameter $\beta$, evaluating the scale range setting $\beta \in [0.3, 1.8]$. We find that the model achieves the best performance at $\beta=1.0$. For the video-question answering task, we assess the scale range settings $\alpha \in [0.5, 3.5]$ (Fig.~\ref{exp: fig1}(c)) and $\beta \in [0.5, 3.5]$ (Fig.~\ref{exp: fig1}(d)). We find that the model achieves the best performance at $\alpha=2.0$ and $\beta=1.0$. Therefore, we establish $\alpha$ as 2.0 and $\beta$ as 1.0 for video-question answering. For the video captioning task, as shown in Fig.~\ref{exp: fig1}(e) and (f), we find that the video captioning task is robust to hyper-parameters $\alpha$ and $\beta$. We consider that this is because we trained the video captioning task for 50 epochs, significantly more than the 5 epochs used for the retrieval and video-question answering tasks. The additional training epochs help mitigate the impact of hyper-parameters $\alpha$ and $\beta$ on model performance. We set $\alpha$ to 1.0 and $\beta$ to 1.0 for video captioning. For the hyper-parameter $\lambda$, as shown in Fig.~\ref{exp: fig1}(g) and (h),  we adopt $\lambda = 2.5$ for video-question answering and $\lambda = 3.3$ for video captioning to achieve optimal performance.

\myparagraph{Effect of the Prediction Header of $\mathcal{R}$.}\label{prediction header} 
Due to the disparity in semantic similarity and interaction index, we design a prediction header to predict the fine-grained relationship $\mathcal{R}_{i,j}$ between the $i_{th}$ video frame and the $j_{th}$ text word. To explore the impact of the structure of the prediction header on our method, we compare four popular structures, \ie ``MLP'', ``CNN'', ``MLP+SA'', and ``CNN+SA''. As shown in Table~\ref{tab: header}, we find that the combination of CNN and attention (``CNN+SA'') captures both local and global interactions, so it is beneficial for predicting the fine-grained relationship between video and text. As a result, we adopt ``CNN+SA'' to achieve optimal performance in practice.

\myparagraph{Effect of the Number of Clusters.}
The cluster module shown in Fig.~\ref{method: fig3} merges the visual tokens and textual tokens, thereby reducing computational overhead. We investigate the impact of merging clusters, as summarized in Table~\ref{tab: abl_cluster}, where $N_{v}^a$ and $N_{v}^o$ represent the number of visual clusters at the action and event levels, respectively. $N_{t}^a$ and $N_{t}^o$ denote the number of text clusters at the action and event levels, respectively. We find the cluster number is a trade-off hyper-parameter: (a) Greater numbers of clusters may lead to semantically similar tokens being classified into different clusters, while (b) fewer clusters may result in dissimilar tokens being grouped together. Upon analysis of Table~\ref{tab: abl_cluster}, we select $\{N^a_v, N^o_v, N^a_t, N^o_t\}$ as $\{6, 2, 16, 4\}$ to achieve optimal performance in practice.

\begin{figure*}
\centering
\includegraphics[width=1.\linewidth]{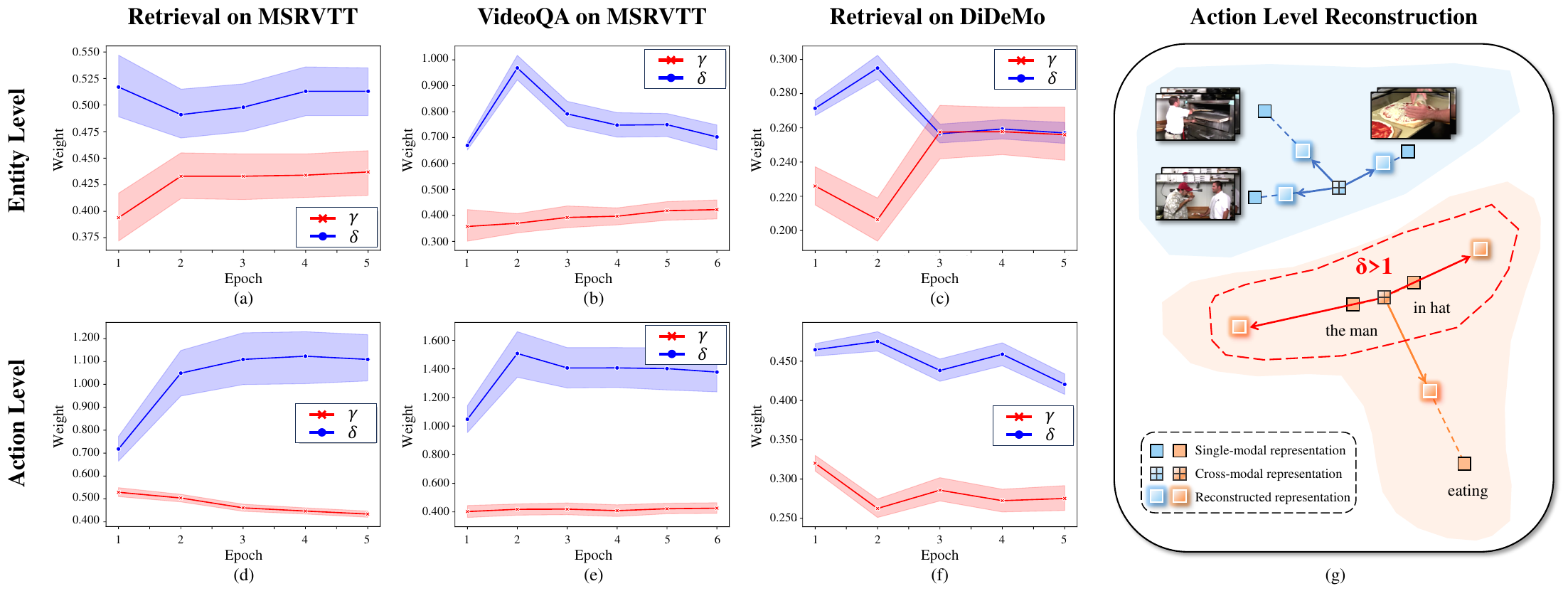}
\vspace{-0.28in}
\caption{\textbf{The convergence curves of learnable video weight $\gamma$ and text weight $\delta$ for both the text-to-video retrieval and VideoQA tasks.} The video weight $\gamma$ stably converges across both entity and action levels, while the text weight $\delta$ initially fluctuates and converges to different weights at different reconstruction levels.}
\label{exp: fig2}
\vspace{-0.1in}
\end{figure*}

\begin{figure*}
\centering
\includegraphics[width=1.\linewidth]{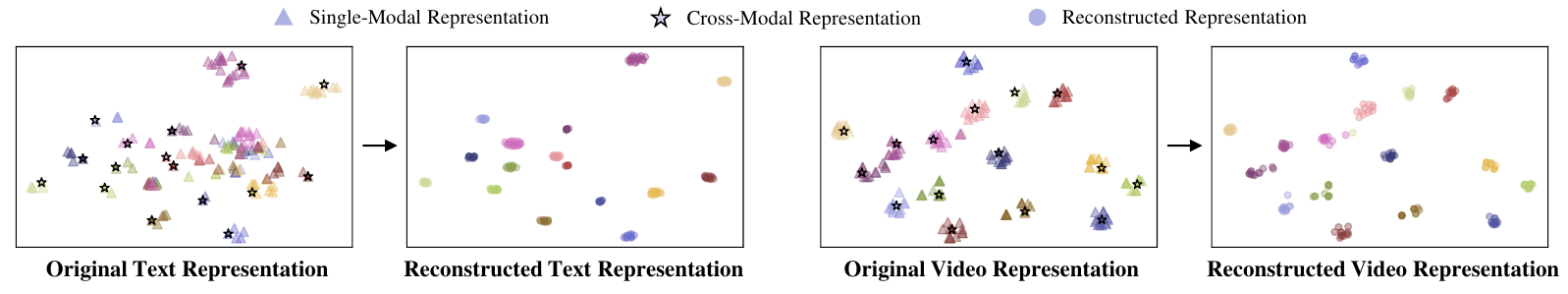}
\vspace{-1.8em}
\caption{\textbf{Visualization (t-SNE~\cite{van2008visualizing}) of the distribution of original and reconstructed representations on the MSRVTT.} We use consistent colors to denote tokens originating from the same text and video. The reconstructed representation maintains the distinctiveness of cross-modal representations while preserving detailed word and frame-level representations.}
\label{fig: visualization of representations}
\end{figure*}

\begin{table*}[t]
\begin{minipage}[c]{0.35\textwidth}
{
    \caption{\textbf{Ablation study for the cluster token number on MSRVTT dataset.} $N_{\_}^a$ and $N_{\_}^o$ denote the number of clusters in action level and event level, respectively.}
    \label{tab: abl_cluster}
    \vspace{-.25em}
    \centering
    \setlength{\tabcolsep}{1.4pt}
    \begin{tabular}{ccccccccc}
        \toprule[1.25pt]
        \multirow{2}{*}{$N^a_v$} & \multirow{2}{*}{$N^o_v$} & \multirow{2}{*}{$N^a_t$} & \multirow{2}{*}{$N^o_t$} & \multicolumn{5}{c}{{Text-to-Video}}\\
        \cmidrule{5-9}
          & & & & R@1$\uparrow$ & R@5$\uparrow$ & R@10$\uparrow$ & MdR$\downarrow$ & MnR$\downarrow$ \\
         \midrule
        9 & 3 & 18 & 4 & 48.7 & 73.8 & \textbf{83.8} & \textbf{2.0} & 12.8\\
        9 & 3 & 16 & 4 & 48.6 & 74.1 & 82.8 & \textbf{2.0} & 12.9\\
        \rowcolor{aliceblue!60} 6 & 2 & 16 & 4 & \textbf{49.4} & \textbf{74.6} & \textbf{83.8} & \textbf{2.0} & \textbf{12.2}\\
        6 & 2 & 12 & 3 & 48.9 & 74.5 & 83.0 & \textbf{2.0} & 12.7\\
        3 & 2 & 8 & 3  & 48.7 & 74.1 & 82.9 & \textbf{2.0} & 12.8\\
        \bottomrule[1.25pt]
    \end{tabular}
}
\end{minipage}
\hfill
\begin{minipage}[c]{0.35\textwidth}
\caption{\textbf{Ablation study for the effect of self-distillation loss on MSRVTT.} $E,A,O$ denote entity level, action level, and event level. ``E-\textgreater{}A\ +\ E-\textgreater{}O'' performs the best.}
\label{tab: self_distill}
\vspace{-.25em}
\centering
\setlength{\tabcolsep}{1.2pt}
\begin{tabular}{lccccc}
    \toprule[1.25pt]
    \multirow{2}{*}{Method} & \multicolumn{5}{c}{{Text-to-Video}}\\
    \cmidrule{2-6}
      & R@1$\uparrow$ & R@5$\uparrow$ & R@10$\uparrow$ & MdR$\downarrow$ & MnR$\downarrow$ \\
     \midrule
    E-\textgreater{}A & 48.7 & 74.3 & 83.5 & \textbf{2.0} & 12.9\\
    E-\textgreater{}O & 48.5 & 73.5 & 81.6 & \textbf{2.0} & 12.8\\
    A-\textgreater{}O & 48.2 & 74.3 & 83.2 & \textbf{2.0} & 13.0\\
    E-\textgreater{}A +\ A-\textgreater{}O & 48.8 & 74.0 & 82.9 & \textbf{2.0} & 12.8\\
    \rowcolor{aliceblue!60} E-\textgreater{}A\ +\ E-\textgreater{}O & \textbf{49.4} & \textbf{74.6} & \textbf{83.8} & \textbf{2.0} & \textbf{12.2} \\
    \bottomrule[1.25pt]
\end{tabular}
\end{minipage}
\hfill
\begin{minipage}[c]{0.27\textwidth}
\caption{\textbf{Iteration time and inference time on the MSRVTT dataset.} We report the average time using two Tesla V100 GPUs.}
\label{tab: time}
\vspace{-.25em}
\centering
\setlength{\tabcolsep}{4.pt}
\begin{tabular}{lcc}
    \toprule[1.25pt]
    \multirow{2}{*}{Method} & Iteration & Inference \\ 
    & Time$\downarrow$ & Time$\downarrow$ \\
     \midrule
    CLIP4Clip~\cite{luo2022clip4clip} & \textbf{1.63} s & \textbf{16.28} s\\
    EMCL~\cite{jin2022expectation} & 1.72 s & 17.68 s\\
    TS2-Net~\cite{liu2022ts2} & 2.57 s & 19.91 s\\
     \midrule
    Baseline & 2.06 s & 18.06 s\\ 
    \rowcolor{aliceblue!60} Ours  & 3.15 s  & 19.18 s\\
    \bottomrule[1.25pt]
\end{tabular}
\end{minipage}
\end{table*} 

\myparagraph{Necessity of the Self Distillation.}
To illustrate the impact of the self-distillation of our method, we conduct ablation experiments on the MSRVTT dataset in Table~\ref{tab: self_distill}. ``$E$'', ``$A$'', ``$O$'' denote entity level, action level, and event level, respectively. ``$-\textgreater{}$'' indicates the distillation direction. For example, ``$E-\textgreater{}A$'' indicates the distillation from ``$E$'' to ``$A$''. We find that the entity level converges first in the training process. This indicates that higher-level semantic features are merged from lower-level semantic features. When lower-level semantic features do not converge, it is difficult for higher-level semantic features to learn semantic information. Based on this observation, we propose using lower-level semantic features to guide the learning of higher-level semantic features. As shown in Table~\ref{tab: self_distill}, self-distillation improves the generalization ability of the model. Distilling from the entity level to the other two semantic levels achieves the best results. Therefore, we distill the entity-level similarity to the other two semantic levels as default in practice.

\myparagraph{The Computational Efficiency of Our Method.}
In Table~\ref{tab: time}, we calculate the iteration time and inference time on the MSRVTT dataset for the retrieval task. Since the Banzhaf Interaction can be removed during inference, our method only takes an additional 1s for processing the test set~(1k videos and 1k text queries). Notably, our method takes less time than TS2-Net during the inference stage. This result demonstrates the superiority of our efficient design.

\begin{figure*}[tbp]
    \centering
    \includegraphics[width=.98\linewidth]{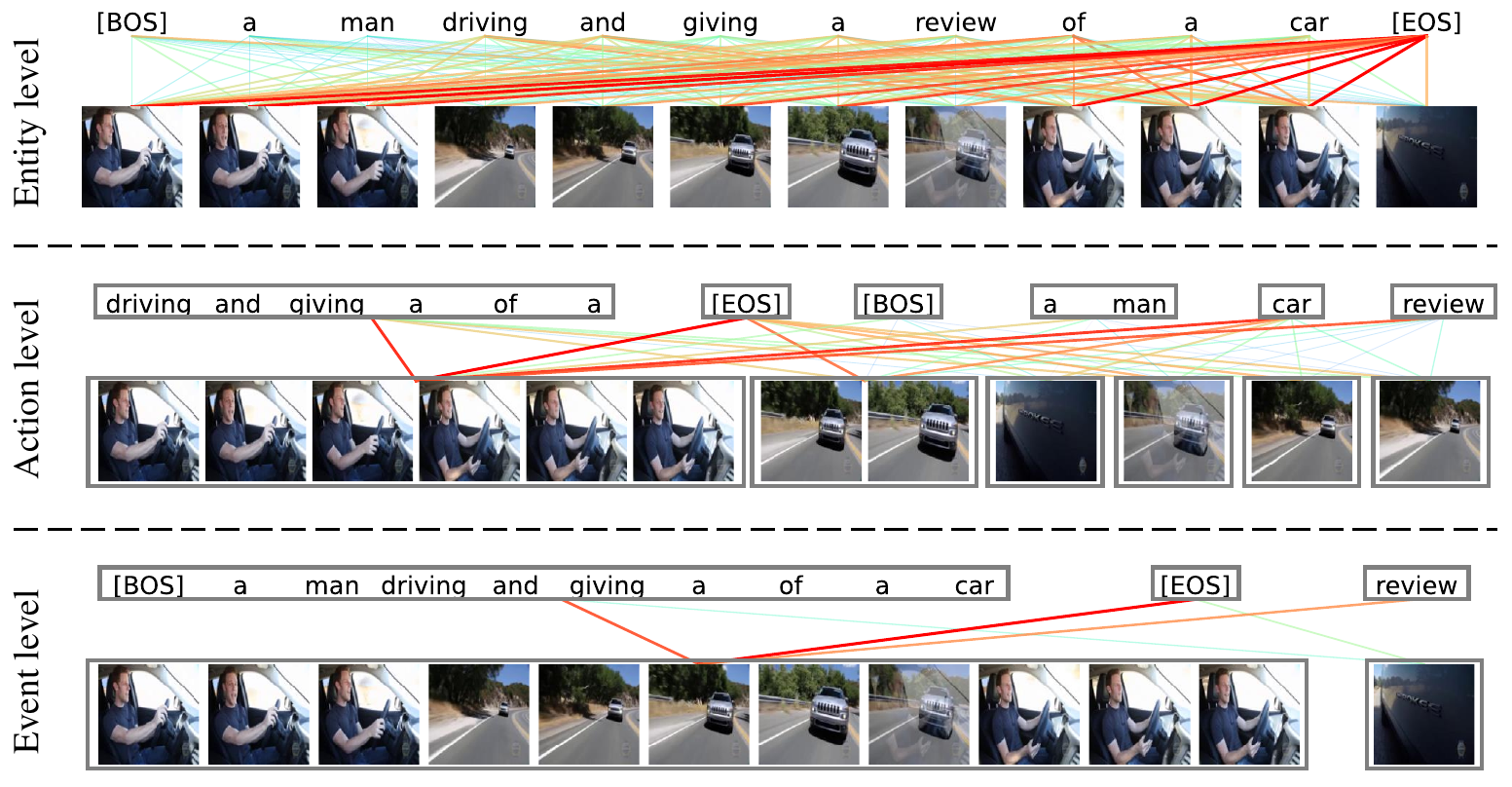}
    \vspace{-.5em}
    \caption{\textbf{Visualization of the hierarchical interaction. We take a representative sample from the MSRVTT dataset as an example.} Here, the degree of confidence from high to low is represented by red, orange, green, and blue lines, respectively.}
    \label{fig: visualization}
    \vspace{-0.10in}
\end{figure*}

\subsection{Discussions}
\myparagraph{Convergence curves of learnable $\gamma$ and $\delta$.}
As shown in Fig.~\ref{exp: fig2}, we observe that the video weight $\gamma$ is more consistent compared to the text weight $\delta$. At both the entity and action levels, $\gamma$ stably converges to 0.4$\sim$0.5 on the MSRVTT dataset and to 0.2$\sim$0.3 on the DiDeMo dataset. This suggests that in video representation, the cross-modal features are beneficial and contribute more significantly than the single-modal features. Although the weight trends are similar in the blue line of Fig.~\ref{exp: fig2}(b) and Fig.~\ref{exp: fig2}(e), the text weight $\delta$ converges to different values across different reconstruction levels. At the entity level, text weight $\delta$ initially fluctuates before stabilizing between 0.7$\sim$0.8, indicating that in entity-level reconstruction for text representation, single-modal information plays a larger role compared to cross-modal. However, at the action level, the text weight $\delta$ fluctuates initially and converges around 1.4 (greater than 1). We believe the model adjusts the weight ratio to make the single-modal information negative, pushing the action-level features away from the central point. This phenomenon is further illustrated in Fig.~\ref{exp: fig2}(g). The degree of clustering among features in the video and text modalities is related to the semantic similarity of the features. At the entity level, there is a significant semantic gap between features in the video and text modalities, with each feature representing different entity information. This noticeable gap helps better match player groups in the game-theoretic interaction. However, the example in Fig.~\ref{exp: fig2}(g) shows that at the action level, the semantic overlap between different text features is much greater, leading to a more concentrated distribution in high-dimensional space, which is actually detrimental to the subsequent game-theoretic interaction. As a result, after incorporating text weights, the model adaptively reduces cross-modal information to widen the feature distribution. In contrast, the action-level video features do not face this issue. We consider this because video frames are semantically richer and naturally more dispersed than captions.

\myparagraph{Visualization of the Representations.}
As shown in Fig.~\ref{fig: visualization of representations}, we find that single-modal representations tend to be widely dispersed and less distinguishable compared to cross-modal representations. However, cross-modal representations typically capture the entire text or video at a coarse-grained level, lacking granularity at the word and frame levels. Our proposed reconstructed representation maintains the distinctiveness of cross-modal representations while also ensuring word and frame-level representations.

\myparagraph{Visualization of the Hierarchical Interaction.}
As shown in Fig.~\ref{fig: visualization}, we find that the semantic similarities between coalitions are generally higher than the semantic similarities between individual frames and individual words. For example, in the action level, the coalition ``\{driving, and, giving, a, of, a\}'' has a high semantic similarity with the video coalition representing the man driving action. On the contrary, when these words interact with the corresponding frame as individuals at the entity level, they show low semantic similarity, while most of the similarity scores are taken by the [EOS] token. This indicates that the model at the entity level learns a coarse-grained alignment between text tokens and video clips. With semantic information, the action level learns a better alignment between text clusters and video clip clusters. The visualization in Fig.~\ref{fig: visualization} also illustrates that our proposed hierarchical interaction method can be used as a tool for visualizing the cross-modal interaction during the cross-modality reasoning tasks. It significantly improves the interpretability of the model and it can help us understand the reasoning details of the models.

\section{Conclusion}
In this work, we creatively model cross-modal representation learning as a multivariate cooperative game by formulating video and text as players in a cooperative game. Specifically, we propose Hierarchical Banzhaf Interaction to value possible correspondence between video frames and text words for sensitive and explainable cross-modal contrast. Although manually labeling the fine-grained relationships between videos and text is unavailable, our method shows a promising alternative to obtaining fine-grained labels based on Banzhaf Interaction. Furthermore, to mitigate the bias in calculations within Banzhaf Interaction, we propose a reconstructed representation that leverages the advantages of both single-modal and cross-modal representations. We extend our original structure into a flexible encoder-decoder framework, allowing the model to adapt to various downstream tasks. To validate the effectiveness and versatility of our method, we conduct extensive experiments encompassing text-video retrieval, video-question answering, and video captioning. More encouragingly, our method can also serve as a visualization tool to promote the understanding of cross-modal interaction.

\section*{Acknowledgement}
This work was supported in part by the National Key R\&D Program of China (No. 2022ZD0118201), the Shenzhen Medical Research Funds in China (No. B2302037), Natural Science Foundation of China (No. 62202014, 62332002, 62425101, 61972217, 32071459, 62176249, 62006133, 62271465), and AI for Science (AI4S)-Preferred Program, Peking University Shenzhen Graduate School, China.

\bibliographystyle{IEEEtran}
\bibliography{egbib}

\vspace{-.4in}
\begin{IEEEbiography}[{\includegraphics[width=0.85in,keepaspectratio]{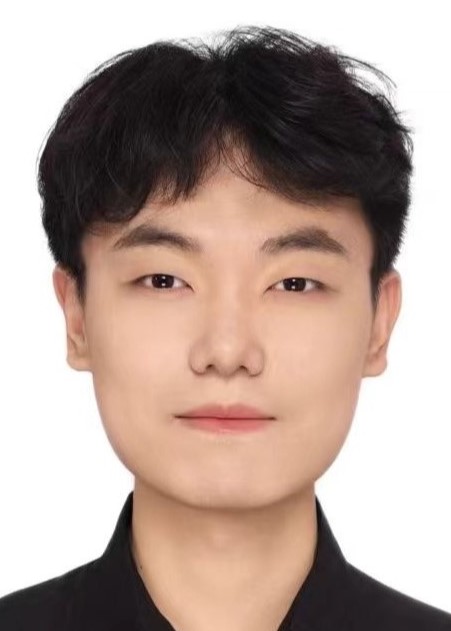}}]{Peng Jin}
received the BE degree in the School of Electronic Information from Tsinghua University, China, in 2021. He is currently working toward a Ph.D. degree at Peking University, Beijing, China. He has published more than 10 papers on top international journals and conferences. His research interests include cross-modality learning and its downstream tasks: multimodal LLMs, video-text retrieval, image caption, and visual question answering. 
\end{IEEEbiography}

\vspace{-.4in}
\begin{IEEEbiography}[{\includegraphics[width=0.85in,clip,keepaspectratio]{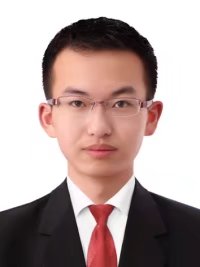}}]{Hao Li}
received the BE degree in the School of Computer Science from Peking University, China, in 2021. He is currently working toward a Ph.D. degree at Peking University, Beijing, China. His research interests include cross-modality learning and its downstream tasks: video-text retrieval, image caption, and visual question answering. 
\end{IEEEbiography}

\vspace{-.4in}
\begin{IEEEbiography}[{\includegraphics[width=0.85in,keepaspectratio]{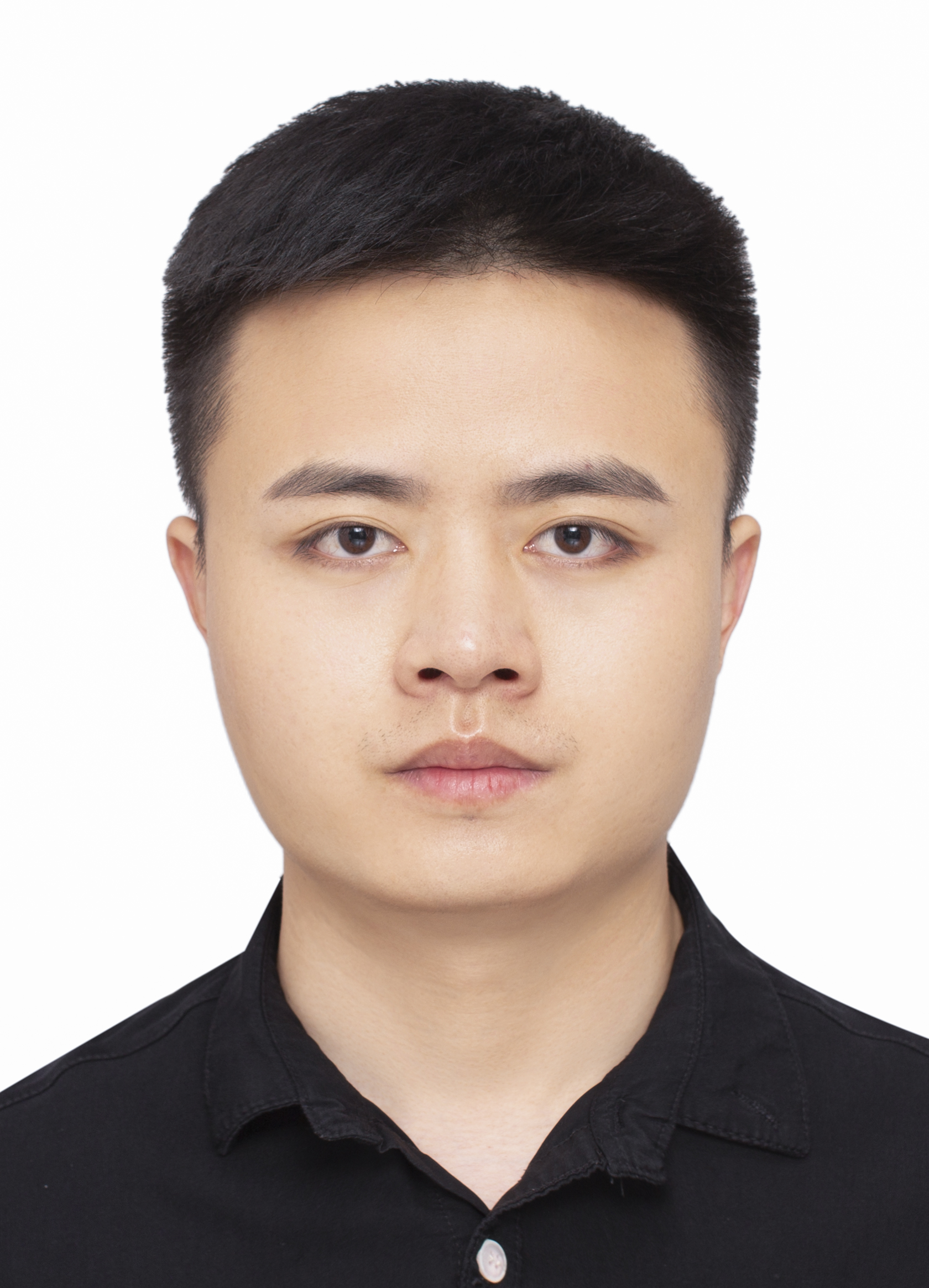}}]
{Li Yuan} received the B.Eng. degree from University of Science and Technology of China in 2017, and Ph.D. degree from National University of Singapore in 2021. He is currently a tenure-track Assistant Professor with School of Electrical and Computer Engineering at Peking University. He has published more than 20 papers on top conferences/journals. His research interests include deep learning, image processing, and computer vision.
\end{IEEEbiography}

\vspace{-.4in}
\begin{IEEEbiography}[{\includegraphics[width=1.0in,keepaspectratio]{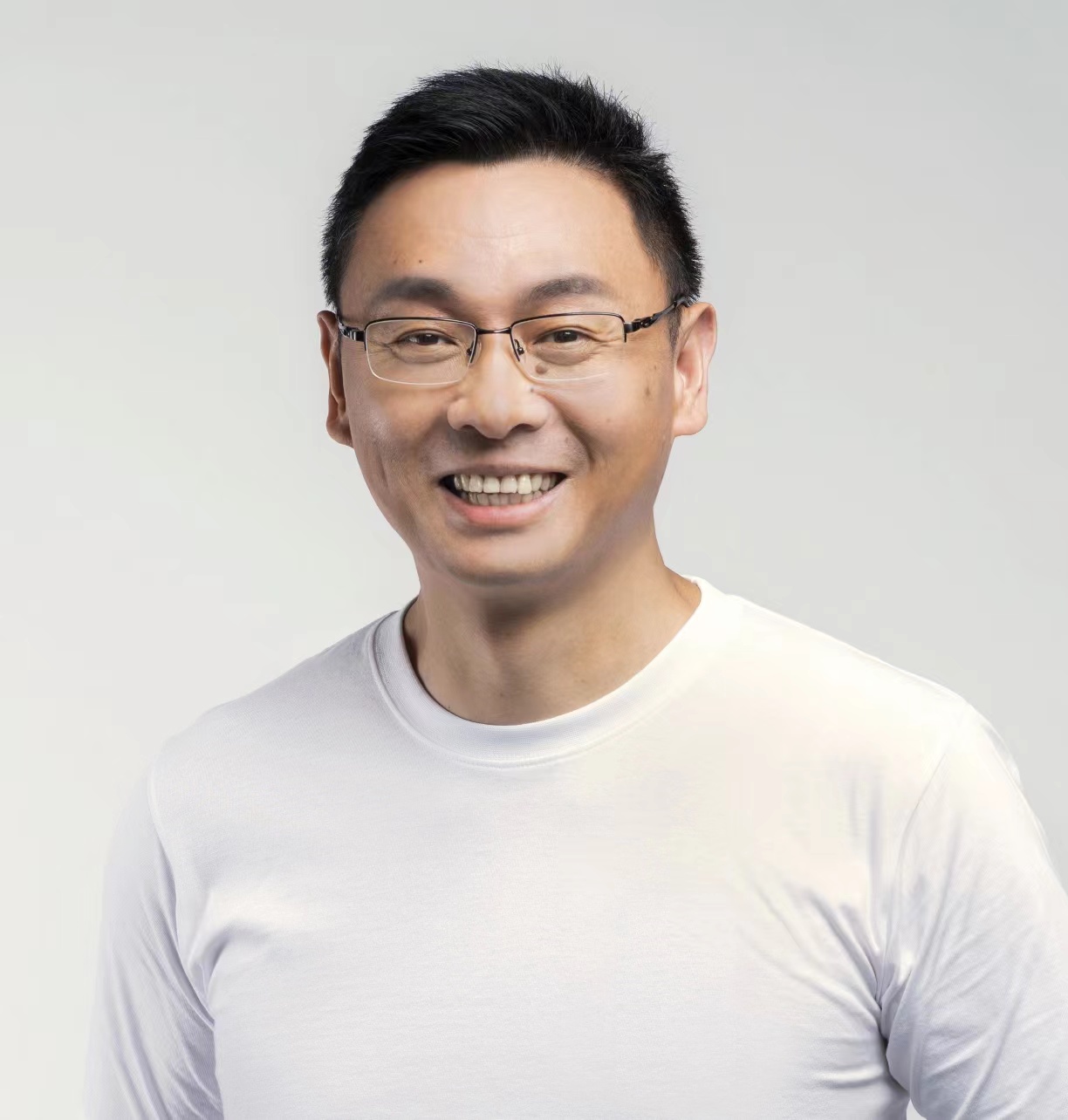}}]
{Shuicheng Yan} (Fellow, IEEE) is currently the Honorary Advisor of Kunlun Tech and was formerly the Chief Scientist at Sea. He is a member of the Singapore Academy of Engineering and a Fellow of AAAI, ACM, IEEE, and IAPR, and his research interests include computer vision, machine learning, and multimedia analytics. To date, Prof. Yan has published more than 800 papers in top international journals and conferences with an H-index of 140+. Prof. Yan Shuicheng's team has won more than ten awards in two core competitions, Pascal VOC and ImageNet (ILSVRC). In addition, his team has won more than ten best paper and best student paper awards, especially a grand slam at ACM Multimedia, the top conference in multimedia, including three best paper awards, two best student paper awards, and a best demo award.
\end{IEEEbiography}

\vspace{-.4in}
\begin{IEEEbiography}[{\includegraphics[width=1.2in,height=1.25in,clip,keepaspectratio]{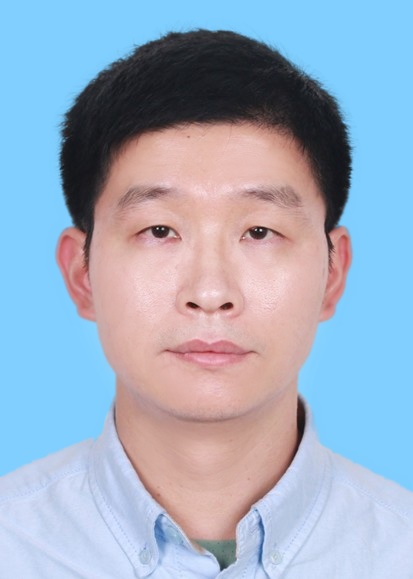}}]{Jie Chen} (Member, IEEE) received the M.Sc. and Ph.D. degrees from the Harbin Institute of Technology, China, in 2002 and 2007, respectively. He joined as the Faculty Member of the Shenzhen Graduate School, Peking University, in 2019. He is currently an Associate Professor with the School of Electronic and Computer Engineering, at Peking University. Since 2018, he has been working with the Pengcheng Laboratory, in China. From 2007 to 2018, he worked as a Senior Researcher with the Center for Machine Vision and Signal Analysis, the University of Oulu, Finland. In 2012 and 2015, he visited the Computer Vision Laboratory, University of Maryland, and School of Electrical and Computer Engineering, Duke University, respectively. He was a Co-Chair of International Workshops at ACCV, ACM MM, CVPR, ICCV, and ECCV. He was a Guest Editor of special issues for IEEE TRANSACTIONS ON PATTERN ANALYSIS AND MACHINE INTELLIGENCE, IJCV, and Neurocomputing. He has been selected as a finalist for the 2022 Gordon Bell Special Prize for HPC-Based COVID-19 Research. His research interests include computer vision, large language model and its application for protein engineering. He has published over 200 papers in top international journals and conferences. He is an Associate Editor of the Visual Computer. 
\end{IEEEbiography}

%\vfill

\end{document}